# Development of Minimal Biorobotic Stealth Distance and Its Application in the Design of Direct-Drive Dragonfly-Inspired Aircraft

Zhang Minghao[a,b,c] Song Bifeng[a,b,c] Yang Xiaojun[a,b,c,*] Wang Liang[a,b,c] Lang Xinyu[a,b,c]

[a] School of Aeronautics, Northwestern Polytechnical University, Xi'an 710072, China;

[b] National Key Laboratory of Aircraft Configuration Design, Xi'an 710072, China;

[c] National Key Laboratory of Strength and Structural Integrity, Xi'an 710072, China;

## Abstract

Advancements in electronic technology and control algorithms have enabled precise flight control techniques, transforming bionic aircraft from principle imitation to comprehensive resemblance. This paper introduces the Minimal Biorobotic Stealth Distance (MBSD), a novel quantitative metric to evaluate the bionic resemblance of biorobotic aircraft. Current technological limitations prevent dragonfly-inspired aircrafts from achieving optimal performance at biological scales. To address these challenges, we use the DDD-1 dragonfly-inspired aircraft, a hover-capable direct-drive aircraft, to explore the impact of the MBSD on aircraft design.

Key contributions of this research include: (1) the establishment of the MBSD as a quantifiable and operable evaluation metric that influences aircraft design, integrating seamlessly with the overall design process and providing a new dimension for optimizing bionic aircraft, balancing mechanical attributes and bionic characteristics; (2) the creation and analysis of a typical aircraft in four directions: essential characteristics of the MBSD, its coupling relationship with existing performance metrics (Longest Hover Duration and Maximum Instantaneous Forward Flight Speed), multi-objective optimization, and application in a typical mission scenario; (3) the construction and validation of a full-system model for the direct-drive dragonfly-inspired aircraft, demonstrating the design model's effectiveness against existing aircraft data.

Detailed calculations of the MBSD consider appearance similarity, dynamic similarity, and environmental similarity. Experimental results indicate that the MBSD value correlates with bionic resemblance and is influenced by design parameters like wingspan, flapping frequency, and amplitude. The study also analyzes the coupling relationships between the MBSD and conventional performance metrics through multi-objective optimization, revealing trade-offs between mechanical performance and bionic resemblance. The final optimized design parameters significantly improve mission performance, demonstrating the practical application of the MBSD in enhancing the operational capabilities of biorobotic aircraft. The MBSD metric proposed in this paper provides valuable guidance for the design and application of related aircraft.

## Keywords

## 1. Introduction

The advancements in electronic technology and control algorithms have enabled precise flight control techniques[1] and direct-drive technology[2], facilitating the evolution of bionic aircraft from mere principle imitation to comprehensive resemblance. This shift has led to bionic aircraft gradually replacing less bionic and more detectable or vulnerable fixed-wing and quad-rotor aircraft in various fields such as patrolling[3], wildlife conservation[4], military reconnaissance[5], and agricultural activities[6]. However, due to current technological limitations, bionic aircraft, especially dragonfly-inspired aircrafts, struggle to achieve optimal performance at biological scales[7-10].

To address the technical challenges faced by bionic aircraft, particularly dragonfly-inspired ones, researchers have been integrating optimization techniques: Hao Zheng et al. [5] optimized biplane FWMAVs using a surrogate model combined with Particle Swarm Optimization, achieving a maximum lift enhancement of 11.5% and at least a 5.4% improvement in torque generation through experimental validations. However, this optimization did not consider constraints or the balance between bionic resemblance and optimization results. Hyeon-Ho Yang et al.[11] optimized foldable flapping-wing aircraft using genetic algorithms to maximize average lift force and minimize total required power, but similarly did not constrain or balance the bionic characteristics. Bosong Duan et al.[12] optimized a bat-like robot using a genetic particle swarm optimization algorithm to minimize flapping energy, constraining the Reynolds number (Re), aspect ratio (AR), Strouhal number (St), and corresponding biological similarities. Yuanbo Dong et al.[13] optimized flapping wings inspired by giant hummingbirds using a surrogate model combined with a Candidate Point Approach, targeting mechanical power while constraining the Reynolds number (Re) and biological similarities.

The analysis of these optimization processes reveals that current bionic aircraft optimization primarily focuses on mechanical performance without adequately balancing bionic and mechanical attributes. Introducing a quantitative metric to evaluate bionic resemblance is crucial to address this issue. This metric can guide the overall system optimization design, allowing a better balance between bionic and mechanical attributes in bionic aircraft.

Regarding evaluating bionic resemblance in bionic aircraft, Lang Tianjiao et al.[14] in 2015 proposed a simulation method for the electromagnetic scattering characteristics of complex multi-flapping bird aircrafts. They extracted flapping frequency based on the simulated radar cross-section (RCS) of bird targets. However, these methods only extracted component motion patterns without analyzing the resemblance. For small bionic aircraft, especially insect-scale ones, the task-specific evaluation of bionic resemblance often relies on visual detection systems like human eyes or high-resolution cameras due to limitations such as electromagnetic equipment blind spots and near-detection distance blind spots. Xiaotong Deng et al.[15] Improved image detection accuracy by 93.1% by embedding spatial and channel attention modules into the RetinaNet detection framework for similar image targets. Bhajan Tri et al.[16] proposed using enhanced texture analysis methods to detect similar image targets. Although these methods achieve high detection accuracy, they primarily focus on differentiation rather than quantitative analysis.

Based on the current research status analysis, this paper proposes a Minimal Biorobotic Stealth Distance evaluation method to assess the bionic level of bionic aircraft, introducing a new research focus. The DDD-1 dragonfly-inspired aircraft is used to explore further the impact of Minimal Biorobotic Stealth Distance on aircraft design. The DDD-1 is a hover-capable direct-drive dragonfly-like aircraft (see Fig. 1) with a take-off weight of 43.5 grams, including a 10.5 grams battery, a wingspan of 26.8 cm, a steady flapping frequency of 23 Hz, and a flapping amplitude of 150 degrees. This aircraft demonstrates hover ability under orbital constraints and attitude stabilization under tethering. Fig. 2(a) shows photos of hover ability under orbital constraints, achieved with an external energy supply since the battery cannot sustain hover. Fig. 2(b) shows photos of attitude stabilization under tethering, enabled due to the motor output's insufficiency for manoeuvring flight power.

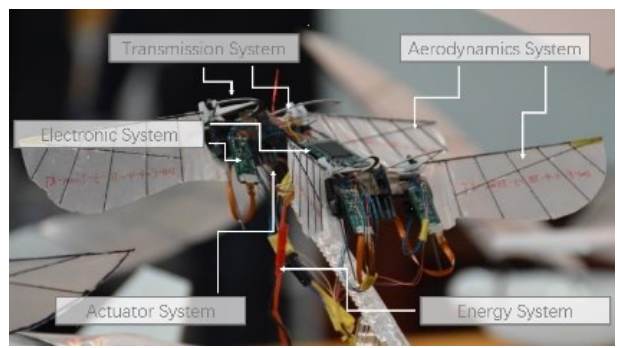

**Fig. 1.** Illustration of system integration of the DDD-1

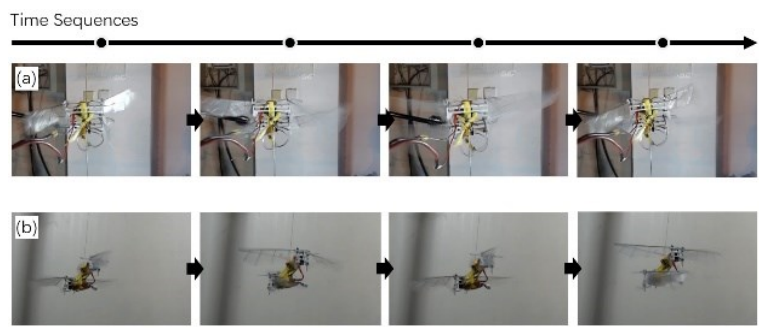

**Fig. 2.** (a) Photos of hovering under orbital constraints under external power supply (b) Photos of

attitude stabilization under tethering

Contributions of this paper are as follows:

1. This paper introduces the Minimal Biorobotic Stealth Distance (MBSD), a quantifiable and operable evaluation metric that influences aircraft design. It integrates seamlessly with the overall design process and provides a new dimension for optimizing bionic aircraft, offering a perspective to balance artificial mechanical attributes and bionic characteristics under current technological conditions.
2. A typical aircrafts established and analyzed in four directions: essential characteristics of Minimal Biorobotic Stealth Distance, its coupling relationship with existing aircraft performance metrics (Longest Hover Duration and Maximum Instantaneous Forward Flight Speed), multi-objective optimization perspective, and optimization in a typical mission scenario using Minimal Biorobotic Stealth Distance.
3. The construction of a full-system model for the direct-drive dragonfly-inspired aircraft and validation of the design model's effectiveness against existing aircraft data.

The remainder of this article is structured as follows: Section 2 provides a detailed definition of Minimal Biorobotic Stealth Distance. Section 3 describes the construction of the aircraft for subsequent analysis of the coupling relationship between Minimal Biorobotic Stealth Distance and existing design objectives. Section 4 discusses the impact of typical design parameters on Minimal Biorobotic Stealth Distance. Section 5 explores Minimal Biorobotic Stealth Distance further by traversing typical design parameters. Section 6 delves into the coupling relationship between Minimal Biorobotic Stealth Distance and existing aircraft design objectives using multi-objective optimization. Section 7 presents a bionic aircraft scenario, optimizing the bionic aircraft using Minimal Biorobotic Stealth Distance as the guiding metric to derive final design parameters.

## 2. Development of Minimal Biorobotic Stealth Distance

The bionic nature of biorobotic aircraft primarily derives from their resemblance to their biological counterparts, allowing them to blend seamlessly into the target species and avoid detection by the naked eye. To develop a metric that balances the mechanical attributes and bionic properties of biorobotic aircraft under current technological conditions, it is essential first to understand the sources of this resemblance. The bionic resemblance of biorobotic aircraft depends on three main aspects:

1. **Appearance Similarity**: The key components or virtual components of the

biorobotic aircraft should resemble the shape of the target biological organism.

2. **Dynamic Similarity**: The motion patterns of the key components of the biorobotic aircraft should mimic those of the target organism, especially in terms of dynamic movement trajectories.

3. **Environmental Similarity**: The overall movement patterns of the biorobotic aircraft should simulate the movement patterns and habits of the target organism. Without this, even if the aircraft achieves 100% appearance and dynamic similarity, it will lack bionic authenticity.

Appearance and environmental similarity can be achieved by developing mimetic shells and task-specific designs, which are facilitated by advancements in manufacturing and coating technologies and biological research[17]. On the other hand, dynamic similarity can be quantified using various parameters[18, 19], with distance being a primary parameter due to the reliance of current biorobotic aircraft on image detectors such as cameras and microphones[20]. Therefore, this paper constructs a detection metric using distance as the parameter, defining the Minimal Biorobotic Stealth Distance.

## 2.1. Calculation Principle of Minimal Visual Resolution Distance

To calculate the Minimal Biorobotic Stealth Distance, it is essential to understand the visual sensor mechanisms of the human eye and cameras. The primary visual sensor for bionic aircraft detection is the human eye. The working mechanism of the human eye involves the crossing of two light rays emitted from a bright point at the eye's nodal point, ultimately focusing on the fovea centralis in the retina. When the images of two points are separated by one cone cell, the human eye can distinguish these two points. This minimal resolvable angle can be determined as follows[21, 22]:

$$\theta_{eye} = arctg\frac{d_1}{D_1} \approx 2.3439'  \qquad (1)$$

Where $d_1$ is the diameter of three cone cells $3 \times 5 \times 10^{-6}m$, and $D_1$ is the distance from the retina to the pupil $2.2 \times 10^{-3}m$. Thus, for a specific detection distance $D$, the minimal resolvable error $d$ for the human eye is:

$$d = D \times \frac{\pi}{180} \times \frac{2.3439}{60} = C_{eye} \times D \qquad (2)$$

$$C_{eye} = \frac{\pi}{180} \times \frac{2.3439}{60} = 6.82 \times 10^{-4} \qquad (3)$$

Based on equation (2), if the resolvable error $d$ is known, the corresponding Minimal Visual Resolution Distance is:

$$D = \frac{d}{C_{eye}} \tag{4}$$

Where $C_{eye}$ is the scaling constant for the human eye.

## 2.2. Mechanism, Influencing Factors, and Construction Principles of Minimal Biorobotic Stealth Distance

To construct the Minimal Biorobotic Stealth Distance, it is essential to consider the working characteristics of the human eye, the dimensions of the biorobotic aircraft, and the correlation between the aircraft's motion and that of the reference biological organism. This paper assumes:

**Assumption 1: Determining Critical Values with the Strictest Standards:** As referenced in the previous calculation of the Minimal Visual Resolution Distance, it is evident that there exists a distance at which two dissimilar objects, when placed far apart, can appear similar. To simplify the influence of observational equipment and enhance the effectiveness of the metric, this paper discusses the Minimal Biorobotic Stealth Distance under the following stringent conditions: the visual detection equipment is directly facing the biorobotic aircraft during the mission, and the detection equipment's ability to discern motion deviations approaches or even exceeds laboratory levels, allowing the identification of all unexpected trajectory deviations. Additionally, it is assumed that the reconnaissance target is fully aware of the motion patterns of the simulated species and can detect any deviations from the expected trajectories. Evidence: Although field conditions are generally less favorable than laboratory conditions, high-precision optical equipment, such as high-resolution telescopes, can still be employed, and field personnel often have extensive experience[23].

**Assumption 2: Using Wingtip Trajectory Error as the Evaluation Metric:** The detection process primarily focuses on the most noticeable parts likely to trigger alerts. In the case of multiple wings, the calculation of the MBSD value should be based on the set of wings with the maximum deviation from the actual biological flyers. This approach excludes the effects of the tail and other moving mechanical parts. Evidence from existing research indicates that the human visual system tends to prioritize the most significant discrepancies[24-26].

## 2.3. Definition of Minimal Biorobotic Stealth Distance

Based on the above discussion and assumptions, the dynamic similarity is calculated by defining the Minimal Biorobotic Stealth Distance. Refer to Fig. 3 for a visual representation. The MBSD is

defined as follows:

$$D_{i,MBSD} = D_{i.shape} + D_{i.trajectory} \quad (5)$$

Where: $D_{i.shape}$ represents the minimum visual resolution distance required to eliminate size dissimilarity. $D_{i.trajectory}$ represents the minimum visual resolution distance required to eliminate motion trajectory dissimilarity.

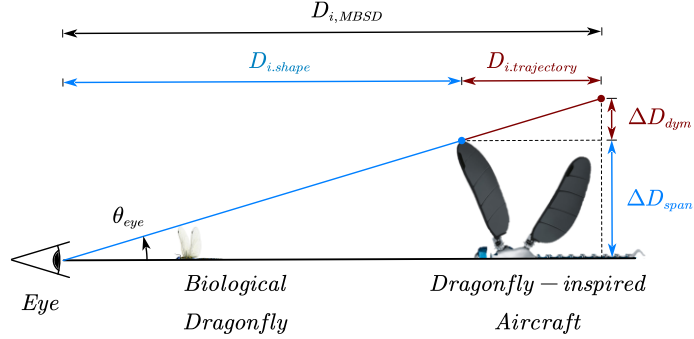

**Fig. 3.** Principle of Minimal Biorobotic Stealth Distance calculation

## 2.4. Calculation of Minimum Visual Resolution Distance for Size Dissimilarity

The minimum visual resolution distance for size dissimilarity is calculated based on the wingspan deviation between the biorobotic aircraft and the biological dragonfly, as shown in equations (6) and (7). Notably, if the wingspan of the biorobotic aircraft is smaller than that of a typical dragonfly[27], it can be considered as blending into the "background noise" of the target species:

$$D_{i.shape} = \frac{\Delta D_{span}}{C_{eye}} \quad (6)$$

$$\Delta D_{span} = \begin{cases} S_{aircraft} - S_{animal} & S_{aircraft} > S_{animal} \\ 0 & S_{aircraft} \leq S_{animal} \end{cases} \quad (7)$$

Where $S_{aircraft}$ is the wingspan of the aircraft, and $S_{animal}$ is the typical wingspan of the corresponding biological organism.

## 2.5. Calculation of Minimum Visual Resolution Distance for Motion Trajectory Dissimilarity

The minimum visual resolution distance for motion trajectory dissimilarity is determined by the difference in motion patterns between the biorobotic aircraft and the biological organism, as shown in equations (8) and (9). It is essential to consider the individual variability among biological dragonflies. If the motion pattern deviation of the biorobotic aircraft is less than the individual variability of the biological dragonfly, it can blend into the "background noise" of the

target species:

$$D_{i.trajectory} = \Delta D_{dym} = \frac{S_{aircraft} \cdot \Delta C_{i.dynamic}}{C_{eye}} \quad (8)$$

$$\Delta C_{i.dynamic} = max(C_{i.dynamic}(f_{wing}, \phi_{am}) - C_{i.individual}, 0) \quad (9)$$

Where $C_{i.dynamic}$ is the distance deviation of motion normalized by wingspan, and $C_{i.individual}$ is the normalized deviation of the biological dragonfly, calculated by averaging the deviation over several cycles.

## 2.5.1. Calculation of $C_{i.dynamic}$

The primary degrees of freedom for a biological dragonfly include flapping plane angle, flapping angle, wing twist angle, and wing sweep angle[28]. However, the current dragonfly-inspired aircraft mainly considers flapping motion. The calculation of $C_{i.dynamic}$ is based on the flapping angle and flapping plane angle. The steps are as follows: selecting the wingspan, flapping frequency, flapping plane angle, flapping amplitude, and flapping median for the biorobotic aircraft to form the wingtip motion trajectory. The Euclidean distance between the biorobotic and reference trajectories is calculated and averaged over multiple cycles:

$$D_{movement} = \text{mean}_{(j=0 \sim n)}(d_{movement,j}) \quad (10)$$

$$C_{i.dynamic} = \frac{D_{movement}}{S_{aircraft}} \quad (11)$$

Where $d_{movement,i}$ is the Euclidean distance between the wingtip motion trajectories at each time step, and $n$ is the total number of time steps.

## 2.5.2. Calculation of $C_{i.individual}$

In the bionic process, the individual variability of the biological flyer is a significant influencing factor. The calculation of $C_{i.individual}$ is shown in Equation (12) and is carried out as follows: under the same wingspan conditions, all possible motion trajectories are generated by combining the observed motion parameter ranges of $m$ individual biological specimens. The maximum deviation between any two motion trajectories is selected and normalized by the wingspan to obtain $C_{i.individual}$:

$$C_{i.individual} = \frac{max_{(i=0 \sim m)}(C_{i.dynamic})}{S_{aircraft}} \quad (12)$$

The statistical results of the motion parameters for biological dragonflies under different flight modes are shown in Table 1, based on precise observations using high-precision equipment under specific conditions:

**Table 1.** Parameters and individual variability calculations for biological dragonflies under different flight modes

| Flight Mode | Flapping Frequency | Flapping Plane Angle | Flapping Amplitude | Flapping Median | Fore-Hind Wing Phase Difference | Individual Variability |
|---|---|---|---|---|---|---|
| Hovering [29, 30] | 38.8~41 | 0~0 | 60~90 | 0~0 | 180 | 0.627 |
| Climbing [31-33] | 27.2~41.5 | 37~66 | 65.2~94 | -3~22.3 | 76.7~102.3 | 0.763 |
| Turning [31, 33-37] | 33.3~41.7 | 19.3°~80° | 31~90 | *-9~1.5* | 0~74 | 0.693 |
| Forward Flight[29, 31, 35-38] | 24~46 | 19.3°~80° | 50~86 | -10.8~7.3 | 60~90 | 0.642 |

Overall, the value of $C_{i.individual}$ is approximately 0.6.

## 3. Construction of the Aircraft Evaluation System

## 3.1. Aircraft Modeling

Referencing the DDD-1 aircraft presented in Fig. 1, a simplified system model as shown in Fig. 4 was constructed. This aircraft is divided into 13 key components: one airframe, four motors, four springs, and four wings. The rig is rigidly attached to the ground at the $O_G^R$ point; the four motors and springs are interconnected and securely mounted on the airframe; and the four wings are individually connected to the rig at the $O_{w,1}^R$, $O_{w,2}^R$, $O_{w,3}^R$, and $O_{w,14}^R$ points, respectively, through constraints with the motors. In the construction of the aforementioned system model, due to the experimental rig's symmetry along the $X_G Z_G$ plane, the points $O_{w,1}$ and $O_{w,2}$ are symmetrical relative to the $X_G Z_G$ plane, as are $O_{w,4}$ and $O_{w,3}$.

The aircraft can be abstracted as a system composed of five rigid bodies with eight degrees of freedom: $\phi_{w,1}$, $\theta_{w,1}$, $\phi_{w,2}$, $\theta_{w,2}$, $\phi_{w,3}$, $\theta_{w,3}$, $\phi_{w,4}$, $\theta_{w,4}$. The system is driven by four motors providing torque. Considering the nonlinear unsteady aerodynamic load, this setup constitutes a multi-body nonlinear underactuated system[7].

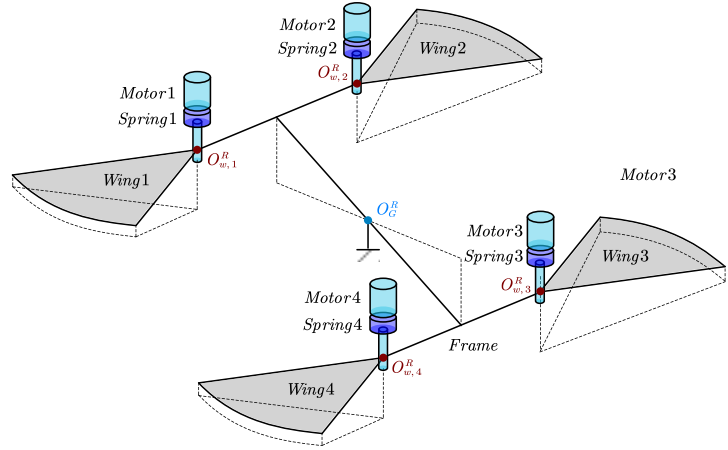

**Fig. 4.** Schematic diagram of the main components of the direct-drive tandem-wing experiment platforms

**Table 2.** Description of the subsystem model

| Subsystem | Function and Calculation method |
|---|---|
| Wing Inertia Model | This module provides the parameters $J_{W,YY}, J_{W,YZ}$, and $J_{W,ZZ}$ required for the simulation. See Appendix A for calculation methods. |
| Torsion Spring Model | This module provides the drive system parameters such as $K_{A,i}$. See Appendix B for calculation methods. |
| Viscoelastic Tensioned Membrane Model | This module provides motion constraints for four torsion angles. Refer to existing studies[41, 42] and Appendix C for calculation methods. |
| Quasi Steady Single Wing Model | This module models the single wing load part of the nonlinear aerodynamic load. Refer to existing studies[43-46] and Appendix D for calculation methods. |
| Tandem Wing Correction Model | This module models the modification term of the tandem wing model relative to the single wing load. Based on the data in the literature[47], see Appendix E for calculation methods. |
| Motor Parameter Database | This module provides parameters such as $I_{max,i}$ for the evaluation process. See Appendix F for calculation methods. |
| Cooling System Model | This module provides parameters such as $P_{HD,i}$ for subsequent constraint analysis. See Appendix G for calculation methods. |
| Trajectory Generator | This module generates the required trajectory forms based on |

| | design parameters. See Appendix H for calculation methods. |
|---|---|
| PID controller | This module provides PID controller to simulate the control process. See Appendix I for calculation methods. |
| First-order Motor and Transmission System Model | This module calculates motor characteristics and transmission system characteristics. See Appendix J for calculation methods. |
| Forward Flight Area Estimation Module | This module estimates the area parameters for calculating forward flight speed. See Appendix K for calculation methods. |
| Battery and Electronic System Model | This module calculates the total battery capacity and other system power consumption. See Appendix L for calculation methods. |

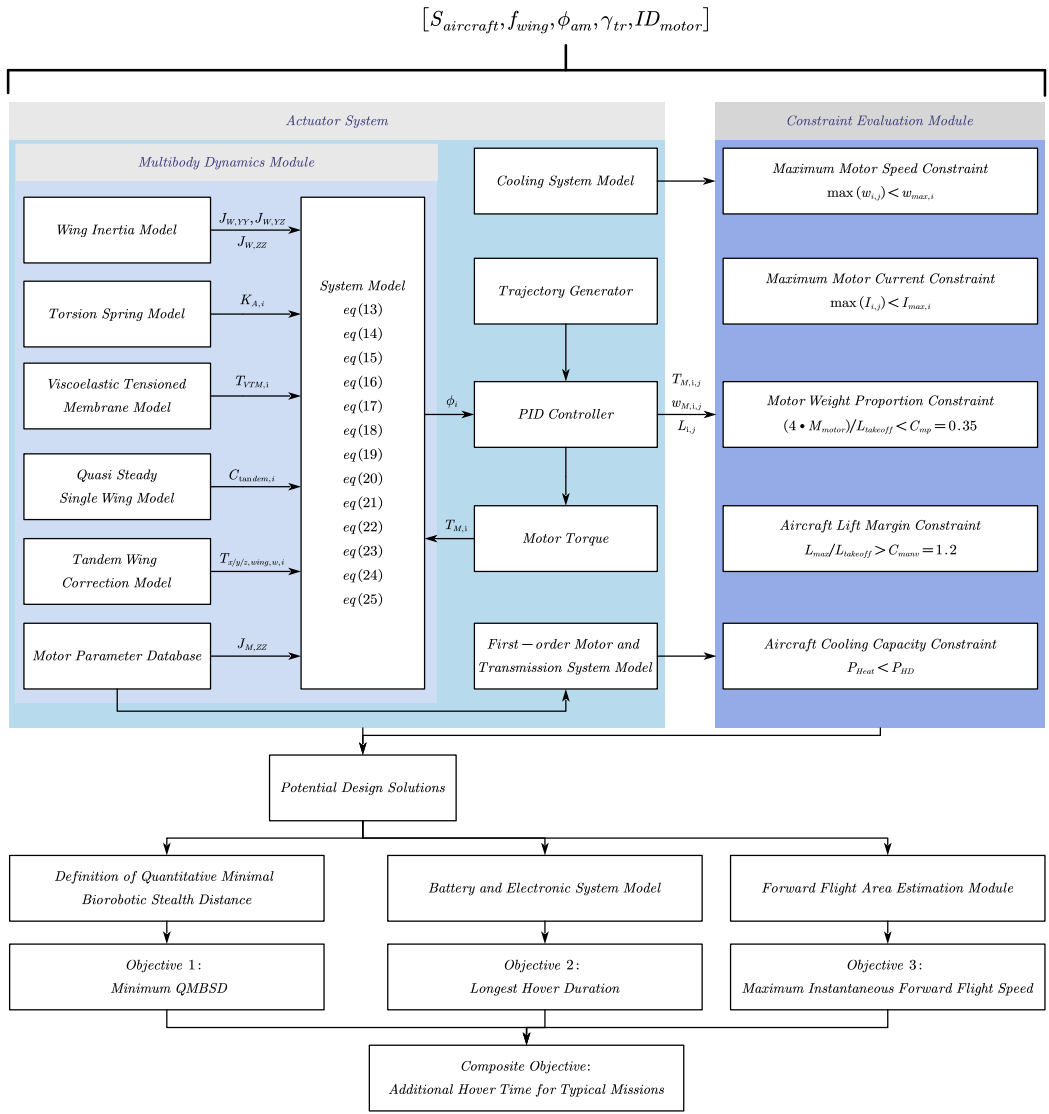

**Fig. 5.** Simulation system and control system framework

Based on the simplified system model of the aircraft and the existing research[7, 39, 40], the modeling approach depicted in Fig. 5 is adopted. The subsystem model in Table 2 is used to obtain the kinetic energy, potential energy, work of generalized forces, and work of nonconservative forces required for the Lagrangian equations, resulting in the system model described by the equations (13) to (25) in the Multibody Dynamics Module. Based on the above, the following 8-degree-of-freedom dynamic equation is obtained using the Lagrangian method[48]:

$$\ddot{\phi}_{w,1} = -\frac{J_{W,YY} \cdot K_{A,1}}{c} \cdot \phi_1 + \frac{J_{W,YY}}{c} \cdot T_{M,1} + \frac{J_{W,YY}}{c} \cdot T_{ZW,1} - \frac{J_{W,YZ}}{c} \cdot T_{YW,1} - \frac{J_{W,YZ}}{c} \cdot T_{VTM,1} \quad (13)$$

$$\ddot{\phi}_{w,2} = -\frac{J_{W,YY} \cdot K_{A,2}}{c} \cdot \phi_2 - \frac{\pi \cdot J_{W,YY} \cdot K_{A,2}}{c} + \frac{J_{W,YY}}{c} \cdot T_{M,2} + \frac{J_{W,YY}}{c} \cdot T_{ZW,2} - \frac{J_{W,YZ}}{c} \cdot T_{YW,2} - \frac{J_{W,YZ}}{c} \cdot T_{VTM,2} \quad (14)$$

$$\ddot{\phi}_{w,3} = -\frac{J_{W,YY} \cdot K_{A,3}}{c} \cdot \phi_3 - \frac{\pi \cdot J_{W,YY} \cdot K_{A,3}}{c} + \frac{J_{W,YY}}{c} \cdot T_{M,3} + \frac{J_{W,YY}}{c} \cdot T_{ZW,3} - \frac{J_{W,YZ}}{c} \cdot T_{YW,3} - \frac{J_{W,YZ}}{c} \cdot T_{VTM,3} \quad (15)$$

$$\ddot{\phi}_{w,4} = -\frac{J_{W,YY} \cdot K_{A,4}}{c} \cdot \phi_4 + \frac{J_{W,YY}}{c} \cdot T_{M,4} + \frac{J_{W,YY}}{c} \cdot T_{ZW,4} - \frac{J_{W,YZ}}{c} \cdot T_{YW,4} - \frac{J_{W,YZ}}{c} \cdot T_{VTM,4} \quad (16)$$

$$\ddot{\theta}_{w,1} = \frac{J_{M,ZZ}}{c} \cdot T_{VTM,1} + \frac{J_{M,ZZ}}{c} \cdot T_{YW,1} + \frac{J_{W,YZ}}{c} \cdot K_{A,1} \cdot \phi_1 - \frac{J_{W,YZ}}{c} \cdot T_{M,1} - \frac{J_{W,YZ}}{c} \cdot T_{ZW,1} + \frac{J_{W,ZZ}}{c} \cdot T_{VTM,1} + \frac{J_{W,ZZ}}{c} \cdot T_{YW,1} \quad (17)$$

$$\ddot{\theta}_{w,2} = \frac{J_{M,ZZ}}{c} \cdot T_{VTM,2} + \frac{J_{M,ZZ}}{c} \cdot T_{YW,2} + \frac{J_{W,YZ} \cdot K_{A,2}}{c} \cdot \phi_2 + \frac{\pi \cdot J_{W,YZ} \cdot K_{A,2}}{c} - \frac{J_{W,YZ}}{c} \cdot T_{M,2} - \frac{J_{W,YZ}}{c} \cdot T_{ZW,2} + \frac{J_{W,ZZ}}{c} \cdot T_{VTM,2} + \frac{J_{W,ZZ}}{c} \cdot T_{YW,2} \quad (18)$$

$$\ddot{\theta}_{w,3} = \frac{J_{M,ZZ}}{c} \cdot T_{VTM,3} + \frac{J_{M,ZZ}}{c} \cdot T_{YW,3} + \frac{J_{W,YZ} \cdot K_{A,3}}{c} \cdot \phi_3 + \frac{\pi \cdot J_{W,YZ} \cdot K_{A,3}}{c} - \frac{J_{W,YZ}}{c} \cdot T_{M,3} - \frac{J_{W,YZ}}{c} \cdot T_{ZW,3} + \frac{J_{W,ZZ}}{c} \cdot T_{VTM,3} + \frac{J_{W,ZZ}}{c} \cdot T_{YW,3} \quad (19)$$

$$\ddot{\theta}_{w,4} = \frac{J_{MZZ}}{c} \cdot T_{VTM,4} + \frac{J_{M,ZZ}}{c} \cdot T_{YW,4} + \frac{J_{W,YZ}}{c} \cdot K_{A,4} \cdot \phi_4 - \frac{J_{W,YZ}}{c} \cdot T_{M,4} - \frac{J_{W,YZ}}{c} \cdot T_{ZW,4} + \frac{J_{W,ZZ}}{c} \cdot T_{VTM,4} + \frac{J_{W,ZZ}}{c} \cdot T_{YW,4} \quad (20)$$

$$c = J_{M,ZZ} \cdot J_{W,YY} + J_{W,YY} \cdot J_{W,ZZ} - J_{W,YZ}^2 \quad (21)$$

$$\begin{bmatrix} 0 \\ T_{YW,1} \\ T_{ZW,1} \end{bmatrix} = \begin{bmatrix} 0 \\ T_{y,\text{wing},w,1} \\ T_{z,\text{wing},w,1} \end{bmatrix} \cdot (C_{tandem,1} + 1) \quad (22)$$

$$\begin{bmatrix} 0 \\ T_{YW,2} \\ T_{ZW,2} \end{bmatrix} = \begin{bmatrix} 0 \\ T_{y,\text{wing},w,2} \\ T_{z,\text{wing},w,2} \end{bmatrix} \cdot (C_{tandem,2} + 1) \quad (23)$$

$$\begin{bmatrix} 0 \\ T_{YW,3} \\ T_{ZW,3} \end{bmatrix} = \begin{bmatrix} 0 \\ T_{y,\text{wing},w,3} \\ T_{z,\text{wing},w,3} \end{bmatrix} \cdot (C_{tandem,3} + 1) \quad (24)$$

$$\begin{bmatrix} 0 \\ T_{YW,4} \\ T_{ZW,4} \end{bmatrix} = \begin{bmatrix} 0 \\ T_{y,\text{wing},w,4} \\ T_{z,\text{wing},w,4} \end{bmatrix} \cdot (C_{tandem,4} + 1) \quad (25)$$

where $J_{W,YY}, J_{W,YZ}$, and $J_{W,ZZ}$ are the wing's inertia parameters. $J_{M,ZZ}$ is the motor's rotational inertia parameter. $K_{A,i}$ is the spring's stiffness coefficient installed at the $i-$th wing's servo motor.

$T_{M,i}$ specifies the output torque of the $i-$ th servo motor. $T_{y,wing,w,i}$ is the torque around the $Y_{w,i}$ axis experienced by the wing surface, generated based on the aerodynamic force of the single wing. $T_{z,wing,w,i}$ is the torque around the $Z_{w,i}$ axis experienced by the wing surface, generated based on the aerodynamic force of the single wing. $C_{tandem,i}$ indicates the influence coefficient of the interference from tandem wings on the $i-$ th wing. $T_{VTM,i}$ is the wing membrane constraint torque of the *i*-th wing.

Referencing Fig. 5, the system framework for the aircraft analysis and optimization process is shown. Based on the system equations given in the Multibody Dynamics Module, the Trajectory Generator, PID Controller, and action modules are introduced to achieve closed-loop control and simulate the operation of the direct-drive system. Additionally, the Cooling System Model and First-Order Motor Model are used to evaluate whether potential design solutions meet design constraints. If the constraints are met, the Definition of Quantitative Minimal Biorobotic Stealth Distance, Battery and Electronic System Model, and Forward Flight Area Estimation Module are used to calculate the three objective performance metrics of the aircraft.

## 3.2. Design Variables

Based on the data from the simulation system construction and existing research on biorobotic aircraft, the following parameters are selected as design variables:

$$[S_{aircraft}, f_{wing}, \phi_{am}, \gamma_{tr}, ID_{motor}] \tag{26}$$

The default values and ranges of these design variables are shown in Table 3:

**Table 3.** Range and default values of design variables selected

| Design Variable | Unit | Default Value | Lower Limit | Upper Limit |
|---|---|---|---|---|
| $\phi_{am}$ | deg | 80 | 10 | 85 |
| $f_{wing}$ | Hz | 34 | 15 | 50 |
| $R$ | m | 0.075 | 0.05 | 0.12 |
| $ID_{motor}$ | / | 3 | 0 | 20 |
| $\gamma_{tr}$ | / | 25 | 5 | 35 |

In this paper, the aspect ratio is set to a constant value of 3.302, and the root-to-tip ratio is set to a constant value of 0.40.

## 3.3. Design Objectives

### 3.3.1. Objective 1: Minimal Biorobotic Stealth Distance

Based on the definition of Minimal Biorobotic Stealth Distance, the calculation is summarized as

follows:

$$D_i = \frac{\Delta D_{span} + S_{aircraft} \cdot \Delta C_{i.dynamic}}{C_{eye}} \tag{27}$$

$$\Delta D_{span} = \begin{cases} S_{aircraft} - S_{animal} & S_{aircraft} > S_{animal} \\ 0 & S_{aircraft} \leq S_{animal} \end{cases} \tag{28}$$

$$\Delta C_{i.dynamic} = max(C_{i.dynamic}(f_{wing}, \phi_{am}) - C_{i.individual}, 0) \tag{29}$$

### 3.3.2. Objective 2: Longest Hover Duration

Hovering capability is a common design metric for biorobotic aircraft[13, 46, 49-52], so the Longest Hover Duration (LHD) is defined as:

$$T_{HOVER} = \frac{E_{total}}{P_{HOVER}} \tag{30}$$

Where $E_{total}$ is the total energy contained in the battery, and $P_{HOVER}$ is the total power consumption during hovering.

### 3.3.3. Objective 3: Maximum Instantaneous Forward Flight Speed

Forward flight speed is another common design metric for biorobotic aircraft[53-56]. Based on the statistical data of dragonfly flight modes in Table 1 and referencing existing aircraft like the Festo Bionicopter[57], the aircraft achieves forward flight by adjusting the flapping plane angle and amplitude. According to equation (31), the magnitude of $T_{rest}$ is the projection of $L_{max}$ in the horizontal plane minus the vertical component used to maintain the aircraft's attitude, i.e.,

$$T_{rest} = \sqrt{L_{max}^2 - L_{hover}^2} \tag{31}$$

Where $L_{max}$ is the maximum lift produced by the aircraft. Both increased frequency and amplitude can produce additional lift. For the given aircraft, the flapping frequency is relatively constant due to its resonant structure, so $L_{max}$ is the lift produced at the maximum amplitude (180 degrees).

Referencing existing literature[58] for drag estimation during the transition from hover to forward flight:

$$T_{rest} = F_D = C_D \cdot \frac{1}{2}\rho_{air}V_{front}^2 \cdot A_{front} \tag{32}$$

Where $C_D$ is obtained from Bernet-Clark[59] and Kovac et al l[60], and experimental validation results[60, 61]. Thus, the Maximum Instantaneous Forward Flight Speed(MIFFS) is:

$$V_{front} = \sqrt{\frac{2 \cdot T_{rest}}{C_D \cdot \rho_{air} \cdot A_{front}}} \tag{33}$$

$$A_{front} = A_{LT} \cdot sin(\beta) \tag{34}$$

Where $A_{LT}$ is limited by the size of the electronic system, based on the DDD-1 shown in Fig.

1; $\beta$ is the angle formed by the total lift vector and the vector required to maintain hovering.

## 3.4. Design Constraints

### 3.4.1. Constraint 1: Maximum Motor Speed Constraint

Exceeding the motor's rated speed can lead to damage to the motor's bearings[62]. Therefore, it is constrained as:

$$\max(w_{i,j}) < w_{max,i} \tag{35}$$

Where: $w_{i,j}$ is the speed of the $i$-th motor at the $j$-th time step; $w_{max,i}$ is the maximum speed that the bearings of the $i$-th motor can withstand.

### 3.4.2. Constraint 2: Maximum Motor Current Constraint

Exceeding the motor's rated current can damage its internal electronic components and lead to overheating[63]. Therefore, it is constrained as:

$$\max(I_{i,j}) < I_{max,i} \tag{36}$$

Where: $I_{i,j}$ is the current of the $i$-th motor at the $j$-th time step; $I_{max,i}$ is the maximum current that the $i$-th motor can withstand, obtained from the motor database.

### 3.4.3. Constraint 3: Motor Weight Proportion Constraint

Excessive motor weight proportion can pose significant challenges to the design of other components such as the electronic system and airframe structure[6-9, 46, 64-66]. Based on Table 4, the motors of direct-drive aircraft typically account for about 30% of the take-off weight. The take-off weight is determined based on motor weight:

**Table 4.** Proportion of motor weight in the take-off weight of existing typical direct-drive dragonfly-inspired aircraft.

| Aircraft | Motor Weight (g) | Take-off Weight (g) | Proportion (%) |
|---|---|---|---|
| Purdue Aircraft [67] | 6.0 | 20.4g | 29.4% |
| CAS Aircraft[68] | 10.13 | 32.97 | 30.7% |

$$L_{takeoff} = L_1 + L_2 + L_3 + L_4 \tag{37}$$

$$\frac{4 \cdot M_{motor}}{L_{takeoff}} < C_{mp} = 0.35 \tag{38}$$

Where $L_1$, $L_2$, $L_2$, and $L_4$ are the lifts of the four wings; $M_{motor}$ is the motor weight, obtained from the motor database in Appendix F.

### 3.4.4. Constraint 4: Aircraft Lift Margin Constraint

Insufficient lift margin relative to the hovering state can prevent the aircraft from utilizing excess lift for maneuvering or generating forward thrust[6, 51, 69-73]. Therefore, it is constrained as follows:

$$\frac{L_{max}}{L_{takeoff}} > C_{manv} = 1.2 \tag{39}$$

$$L_{max} = L_{max,1} + L_{max,2} + L_{max,3} + L_{max,4} \tag{40}$$

Where $L_{max,1}$, $L_{max,2}$, $L_{max,3}$, and $L_{max,4}$ are the average lifts at maximum flapping amplitude.

### 3.4.5. Constraint 5: Aircraft Cooling Capacity Constraint

The average heat generation of the motor exceeding the cooling capacity of the drive system can lead to overheating and damage to the motor[74-76]. Therefore, it is constrained as:

$$P_{Heat,i} < P_{HD,i} \tag{41}$$

Where: $P_{Heat,i}$ is the average heat generation of the $i$-th motor; $P_{HD,i}$ is the cooling capacity of the $i$-th motor's drive system.

## 3.5. System Model Accuracy Verification

To verify the accuracy of the system model in Fig. 5 and ensure the effectiveness of the coupling relationship between Minimal Biorobotic Stealth Distance and existing design objectives, the simulation system data are compared with the aircraft data in Table 5:

**Table 5.** Model accuracy validation based on existing direct-drive aircraft data.

| Aircraft | Purdue Aircraft [67] (no tandem wing model) | CAS Ground Experiment [68] | CAS Flight Test [68] |
|---|---|---|---|
| Single Wing Aspect Ratio | 3.302 (measured from image | 3.7 | 3.7 |
| Root-to-Tip Ratio | 0.4 | 0.6 (measured from image | 0.6 (measured from image |
| Flapping Amplitude | 80 | 95.0 | 84.0 |
| Flapping Frequency | 34 | 28.0 | 28.0 |
| ID of Motors | 4 | 3 | 3 |
| Single Wing Lift (g) | 10.200 | 11.000 | 8.243 |
| Calculated Single Wing Lift (g) | 9.345 | 10.035 | 7.908 |
| Accuracy | 8.382% | 8.773% | 4.064% |
| Paper Stated Single Drive System Power Consumption | 5.610 | 5.439 | 3.441 |
| Calculated Single Drive Average Power | 4.730 | 5.620 | 3.888 |
| Accuracy | 15.686% | 3.328% | 12.990% |

The aerodynamic model's accuracy obtained from the quasi-steady model is relatively high. The overall power consumption model is accurate for the CAS ground experiment but less accurate for the aircraft data. Possible reasons include:

1. Key model parameters such as the root-to-tip ratio were inferred from images rather than specified in the paper.
2. Controller parameters in the study may differ from the parameters of the Purdue Aircraft and the CAS aircraft.
3. The aircraft lacks high-precision sensors available in ground experiments, leading to less accurate data recording.
4. The aircraft's load varies dynamically, and the Purdue University aircraft's smaller relative size might be due to maneuvering data included in the flight records, which were not accounted for in the current model due to a lack of specific control commands.
5. Testing conditions may not have been fully documented, as suggested by the larger discrepancy in the CAS data, potentially due to experimental methods introducing unrecorded errors such as support structures, scale measurement accuracy, and unaccounted-for component resistance.

# 4. Analysis of basic characteristics of Minimal Biorobotic Stealth Distance

## 4.1. Analysis of the Impact of Wingspan on Minimal Biorobotic Stealth Distance

According to the calculation given in Equation (28), it is evident that the MBSD value is significantly influenced by the wingspan. To analyze the impact of wingspan on the MBSD value, this study selects a typical half-wingspan range for existing aircraft: 0mm to 500mm. The common yellow dragonfly, with a wingspan of 60mm, was selected as the biological reference, using half of this span $S_{animal} = 30$. The calculations reveal that, as illustrated in Fig. 6, the MBSD value is zero when the biorobotic aircraft's wingspan is within the typical biological wingspan, and it increases linearly when the wingspan exceeds that of the biological flyer.

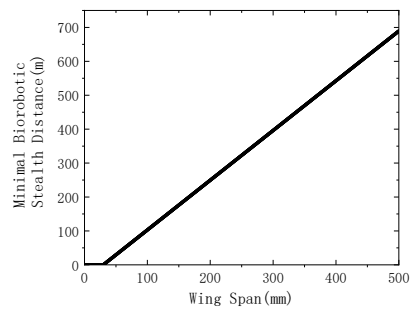

**Fig. 6**. Variation of the MBSD with changes in aircraft wingspan

## 4.2. Analysis of the Impact of Motion Patterns on Minimal Biorobotic Stealth Distance

As indicated by Equation (29), $\Delta C_{i.dynamic}$ is influenced by the coupling effect of $f_{wing}$ and $\phi_{am}$. To study this coupling effect, a range of flapping amplitudes from 20 degrees to 100 degrees and frequencies from 5Hz to 65Hz were selected. The results of these trajectory combinations are shown in Fig. 7:

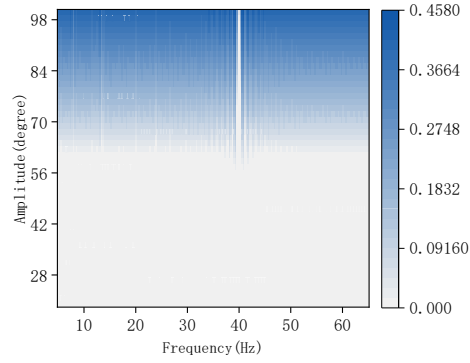

**Fig. 7.** Analysis of the impact of flapping amplitude and frequency on the MBSD

Based on the analysis, the following observations can be made:

1. Near the reference trajectory frequency (39.9Hz), various flapping amplitudes result in smaller MBSD values, indicating better bionic resemblance. This suggests that the system is less sensitive to amplitude variations, providing greater design flexibility in balancing mechanical and bionic attributes.
2. At different frequencies, amplitudes below 120 degrees result in smaller MBSD values, indicating better bionic resemblance. So, an amplitude of 60 degrees can be used as an initial value for subsequent aircraft design.
3. Larger flapping amplitudes generally result in higher MBSD values, indicating a significant decrease in bionic resemblance.

## 4.3. Calculation of Minimal Biorobotic Stealth Distance for Existing Aircraft

To further discuss the characteristics and validity of the MBSD value, this study selects data from hovering-type dragonfly-inspired aircraft with certain flight capabilities and calculates their MBSD values. The results are shown in Table 6:

**Table 6.** Calculation results of Minimal Biorobotic Stealth Distance for existing typical dragonfly-inspired aircraft.

| Aircraft | $D_{i.shape}$(m) | $D_{i.trajectory}$(m) | MBSD(m) |
|---|---|---|---|
| Festo BionicOpter[57] | 417.9 | 0.0 | 417.9 |
| DragonflEye[77] | 0.0 | 0.0 | 0.0 |
| DElFLY Nimble[78] | 197.9 | 0.0 | 197.9 |
| QV[79] | 66.0 | 0.0 | 66.0 |
| DDD-1 | 102.6 | 10.3 | 112.9 |

From the calculations, the following conclusions can be drawn:

1. The MBSD value used to evaluate the bionic resemblance of aircraft aligns with

intuitive perceptions of different aircraft.

2. The DragonflEye[77], a modified biological dragonfly, has an MBSD value of 0, indicating that it cannot be distinguished from a real dragonfly even at very close distances, thus exhibiting an extremely high bionic resemblance.

3. Due to its larger size, The Festo BionicOpter[57] is less likely to be perceived as similar to a real dragonfly at close range.

4. Among the three aircraft: DElFLY Nimble[78], QV[79], and DDD-1—the QV[79] shows the best bionic resemblance due to its better size and trajectory details. Although DElFLY Nimble[78] also has four wings, its wingspan and motion trajectory differ significantly from those of a biological dragonfly, making it less likely to be perceived as a real dragonfly.

5. The trajectory-related distance needs to consider both wingspan and structural form. Although the trajectories of DDD-1 and Festo BionicOpter[57] differ $D_{i.trajectory}$, the larger wingspan of Festo BionicOpter[57] results in a higher $D_{i.shape}$, leading to a higher overall MBSD value. This indicates that for certain biological flyers like dragonflies, achieving high degrees of freedom through complex mechanical systems does not necessarily enhance overall bionic resemblance. Using simpler mechanical forms under similar technological conditions has the potential to achieve higher overall bionic resemblance.

# 5. Analysis of the Coupling Relationship Between Minimal Biorobotic Stealth Distance and Typical Aircraft Indicators from the Perspective of Key Parameter Traversal

## 5.1. Experiment Setup

Typical design objectives in the current aircraft design process include Longest Hover Duration and Maximum Instantaneous Forward Flight Speed. To analyze the coupling relationship between Minimal Biorobotic Stealth Distance and these design objectives, this study references existing research and selects frequency, wingspan, and amplitude as key design parameters[11, 13, 46, 80-85]. By traversing these design parameters, this study demonstrates the balance between the bionic attributes represented by the MBSD and the artificial mechanical attributes represented by the other two objectives under the current technological conditions of biorobotic aircraft.

In this section, key parameters are traversed. The design parameters and their discrete forms are selected as shown in Table 7, with other variables set to their default values.

**Table 7.** Value ranges of key parameters in this section.

| Parameter | Unit | Range | Discrete | Reason for Discrete Quantity Setting |
|---|---|---|---|---|
| $f_{wing}$ | Hz | [15,50] | 70 | To achieve a discrete precision of 0.5 Hz |
| $S_{aircraft}$ | mm | [50,120] | 28 | To achieve a discrete precision of 2.5 mm |
| $\phi_{am}$ | deg | [10,85] | 38 | To achieve a discrete precision of 2 degrees |

### 5.1.1. Calculation Method of Normalized Combination Result

To clearly understand how two parameters are coupled as key parameters change, the following normalized relationship is defined:

$$y = f_{normalize}(v_0, \cdots, v_i, \cdots, v_n) = \frac{v_0, \cdots, v_i, \cdots, v_n}{\max(v_0, \cdots, v_i, \cdots, v_n)} \quad (42)$$

In considering the coupling relationship between two indicators, the data for indicators $T_1$ and $T_2$ are processed to obtain the normalized result $C_{value}$:

$$C_{value} = f_{normalize}((1 - w_1) \cdot f_{normalize}(T_1) + (w_1) \cdot f_{normalize}(T_2)) \quad (43)$$

Where $w_1$ is the weight distribution between the two objectives, $T_1$ is the value of objective 1, and $T_2$ is the value of objective 2.

### 5.1.2. Inverse of Minimal Biorobotic Stealth Distance

To visualize the parameters more conveniently in this section, the inverse relationship of the MBSD is used:

$$C_{\text{MBSD},i,j} = \frac{1}{\text{MBSD}_{i,j}} \quad (44)$$

Where $\text{MBSD}_{i,j}$ represents the relative position in the two-dimensional parameter space formed by the two design variables.

## 5.2. Coupling Relationship Between Minimal Biorobotic Stealth Distance and Longest Hover Duration under Different Key Design Parameters

To analyze the coupling relationship between Minimal Biorobotic Stealth Distance and Longest Hover Duration in the combined space of flapping frequency and wingspan:

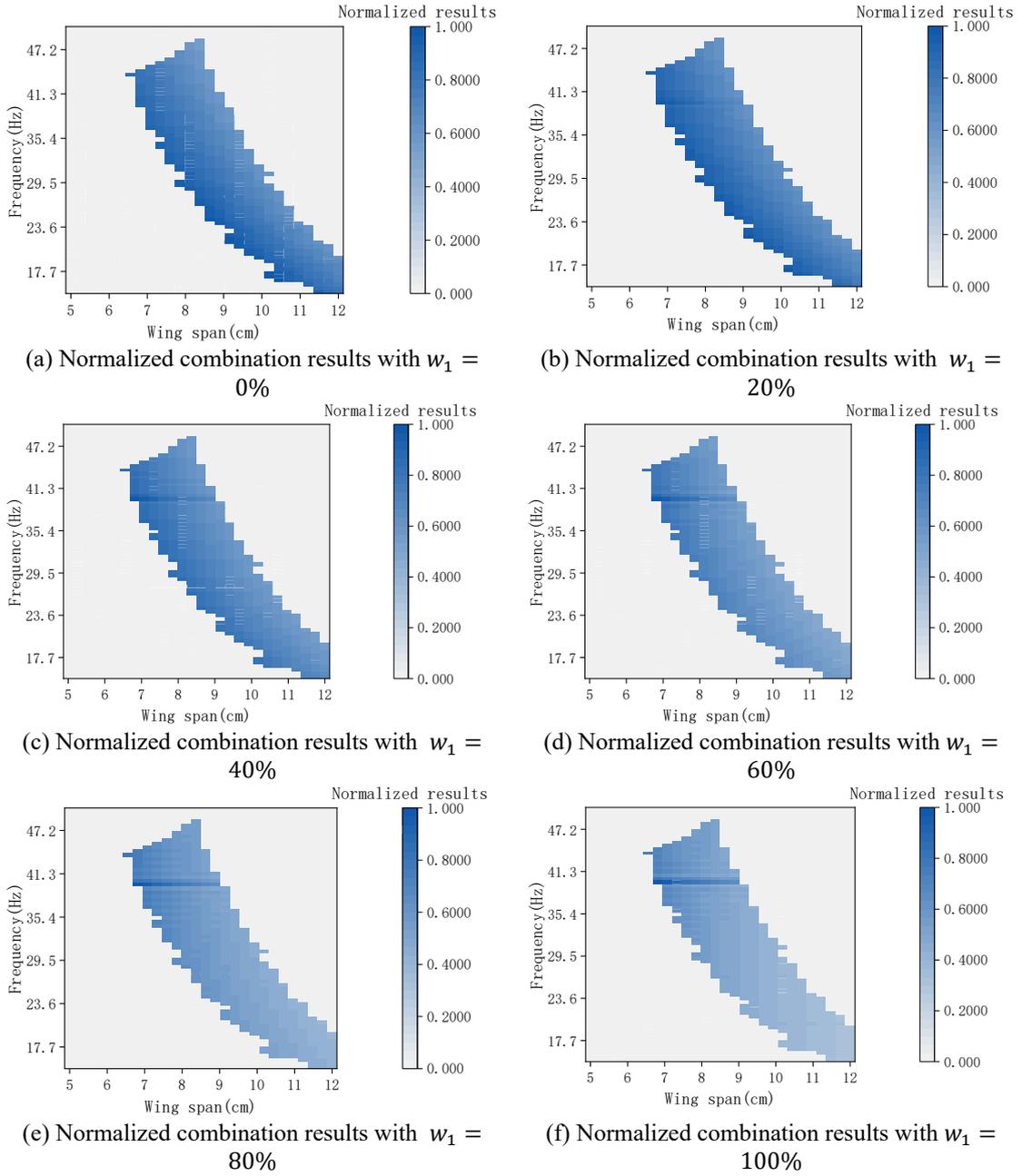

(a) Normalized combination results with $w_1 = 0\%$

(b) Normalized combination results with $w_1 = 20\%$

(c) Normalized combination results with $w_1 = 40\%$

(d) Normalized combination results with $w_1 = 60\%$

(e) Normalized combination results with $w_1 = 80\%$

(f) Normalized combination results with $w_1 = 100\%$

**Fig. 8.** Coupling relationship between the MBSD and the LHD in the design space composed of frequency and wingspan with different weights.

To analyze the coupling relationship between Minimal Biorobotic Stealth Distance and Longest Hover Duration in the combined space of frequency and amplitude:

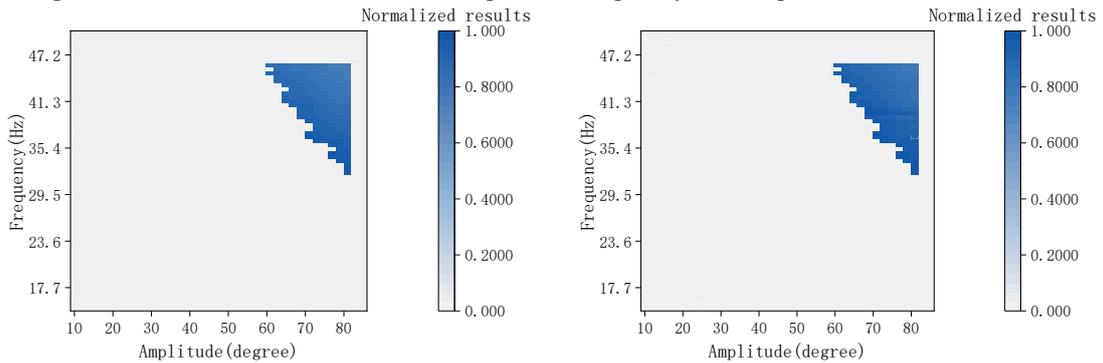

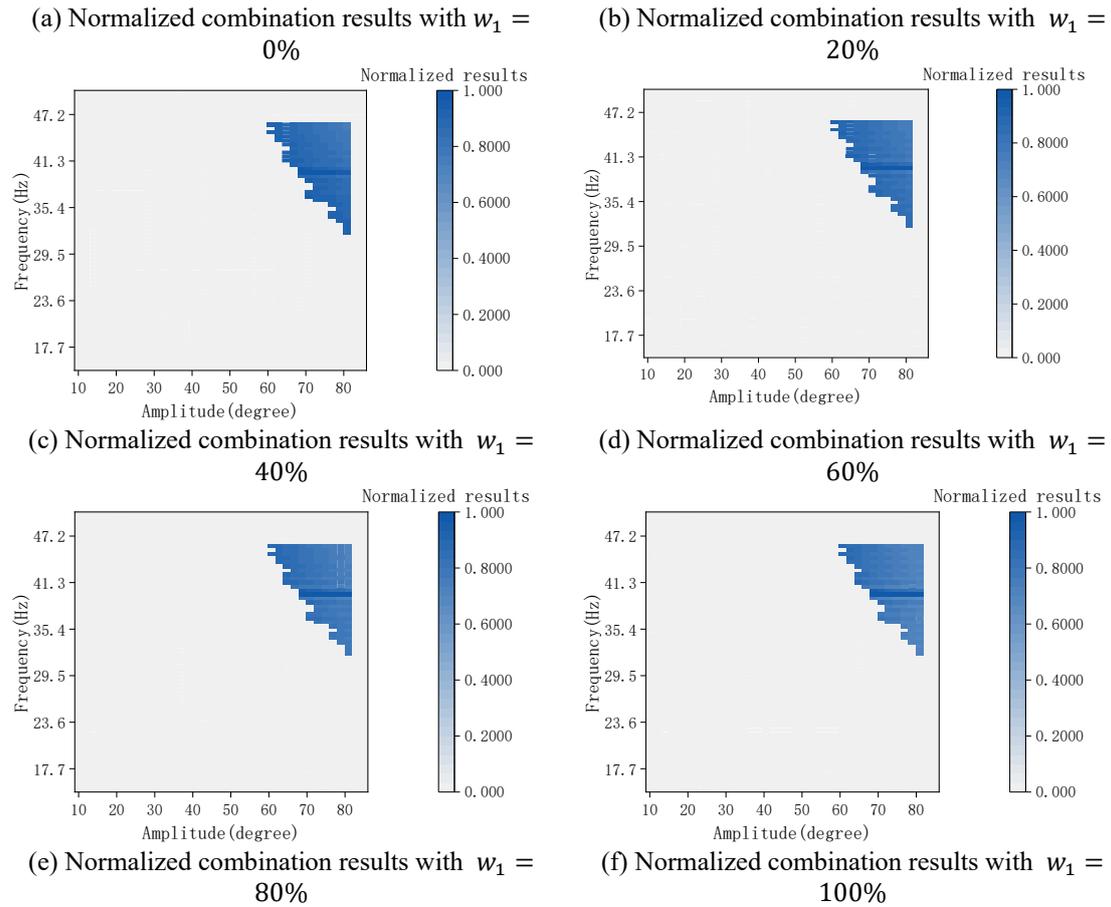

**Fig. 9.** Coupling relationship between the MBSD and the LHD in the design space composed of frequency and amplitude with different weights.

To analyze the coupling relationship between Minimal Biorobotic Stealth Distance and Longest Hover Duration in the combined space of amplitude and wingspan:

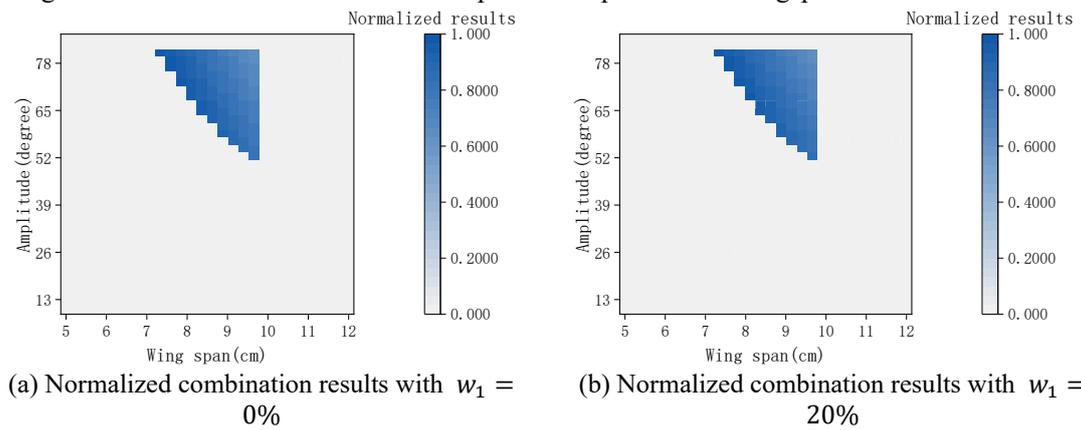

(a) Normalized combination results with $w_1 = 0\%$

(b) Normalized combination results with $w_1 = 20\%$

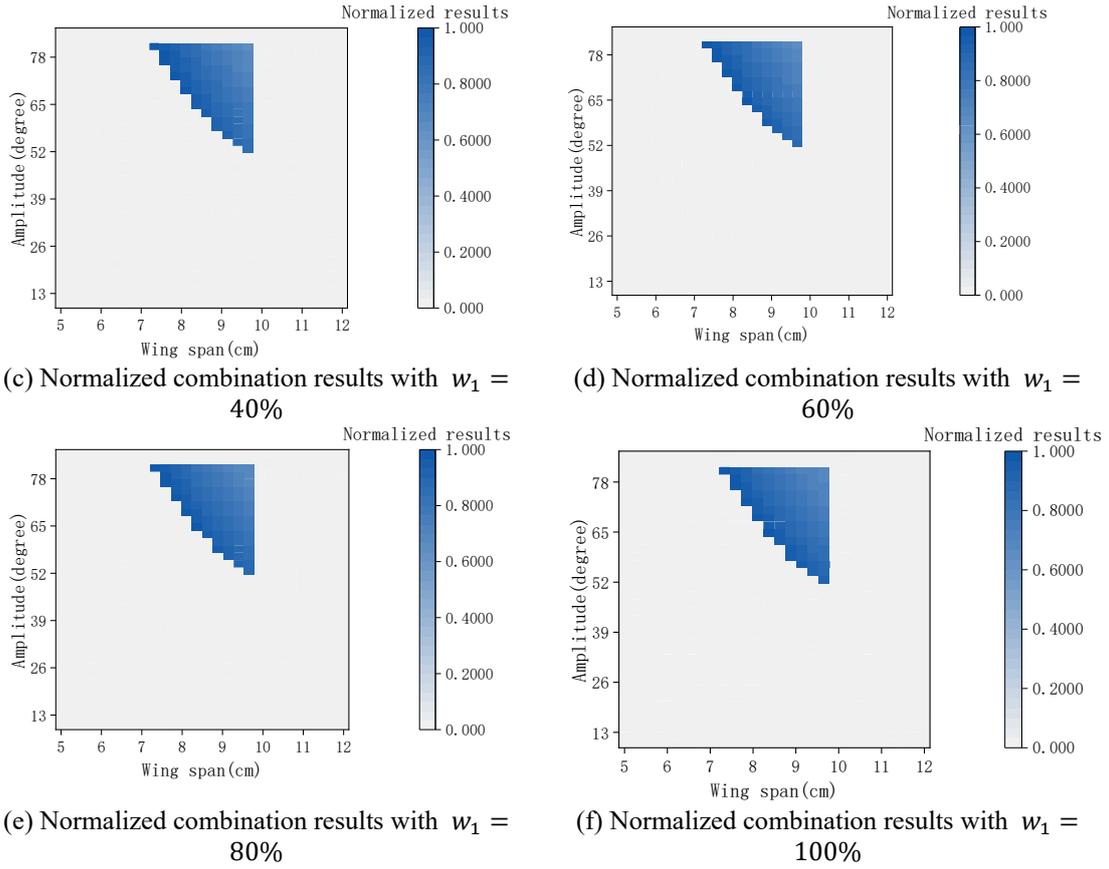

(c) Normalized combination results with $w_1 = 40\%$

(d) Normalized combination results with $w_1 = 60\%$

(e) Normalized combination results with $w_1 = 80\%$

(f) Normalized combination results with $w_1 = 100\%$

**Fig. 10.** Coupling relationship between the MBSD and the LHD in the design space composed of amplitude and wingspan with different weights.

Based on Figs. 8 to 10, the following can be observed:

1. Compared with the distribution of the LHD values, the inverse value of the MBSD tends to achieve larger values (i.e., better bionic resemblance) at a specific frequency in the design parameter space formed by frequency, amplitude, and wingspan. This indicates that selecting an appropriate frequency minimizes the MBSD, highlighting the contradictory nature of the two objectives.

2. As shown in Fig. 8, introducing the MBSD value reveals that in the combined space of the LHD and the MBSD, aircraft with specific frequencies have a greater design advantage over others, highlighting the coupling of the two objectives.

3. The trends influenced by wingspan and amplitude show significant similarities for both objectives. In both cases, the inverse value of the MBSD (i.e., better bionic resemblance) and the LHD value achieve maximum values at smaller wingspans and amplitudes. This suggests that higher bionic resemblance corresponds to higher overall hover efficiency, consistent with existing research[5, 7, 46, 86-89], highlighting the correlation between the objectives.

4. The MBSD value provides a perspective on balancing the mechanical attributes (e.g., the LHD value) and bionic attributes (e.g., the MBSD value) of biorobotic aircraft under current technological conditions.

## 5.3. Coupling Relationship Between Minimal Biorobotic Stealth Distance and Maximum Instantaneous Forward Flight Speed under Different Key Design Parameters

To analyze the coupling relationship between Minimal Biorobotic Stealth Distance and Maximum Instantaneous Forward Flight Speed in the combined space of frequency and wingspan:

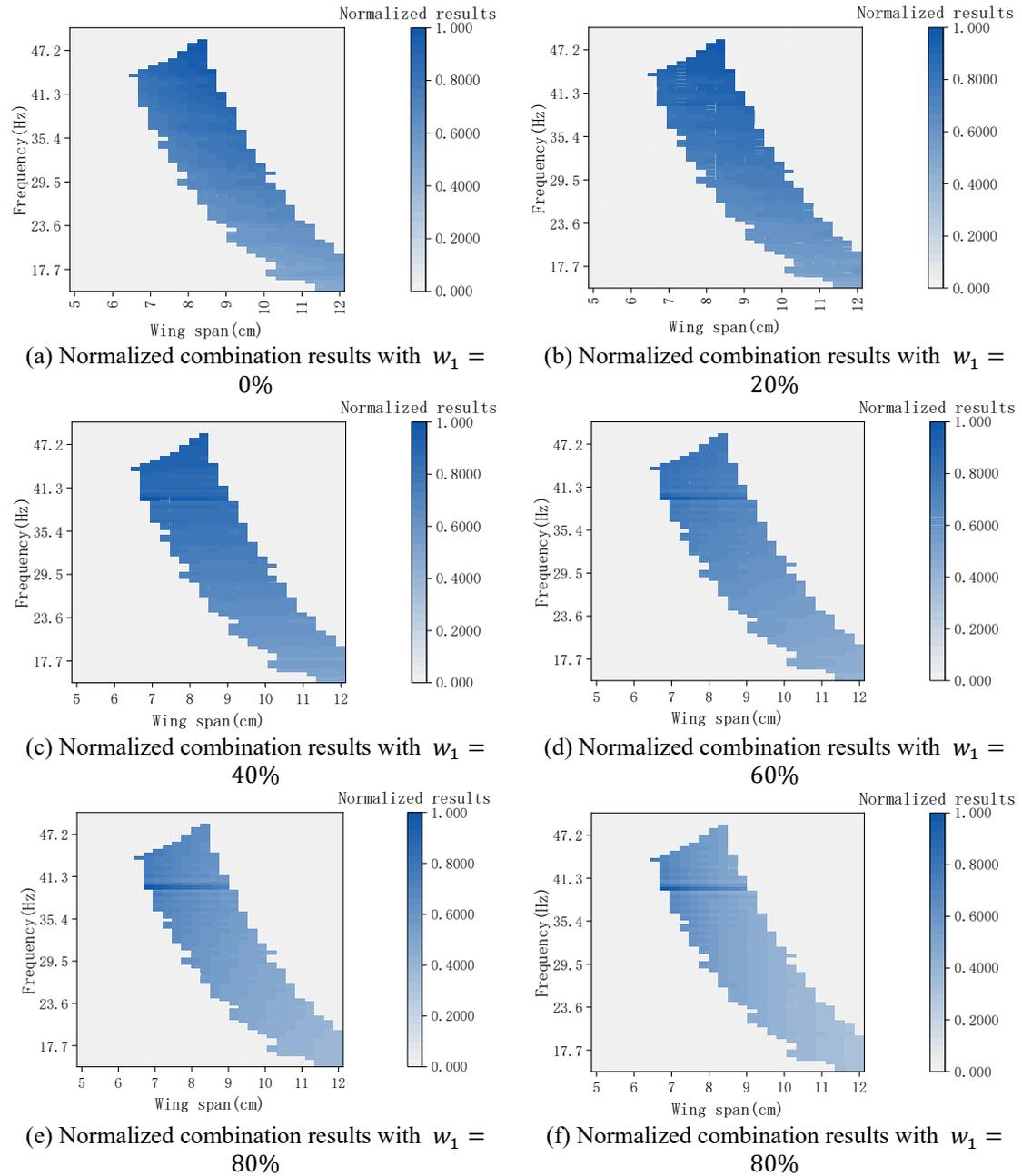

(a) Normalized combination results with $w_1 = 0\%$

(b) Normalized combination results with $w_1 = 20\%$

(c) Normalized combination results with $w_1 = 40\%$

(d) Normalized combination results with $w_1 = 60\%$

(e) Normalized combination results with $w_1 = 80\%$

(f) Normalized combination results with $w_1 = 80\%$

**Fig. 11.** Coupling relationship between the MBSD and the MIFFS in the design space composed of frequency and wingspan with different weights.

To analyze the coupling relationship between Minimal Biorobotic Stealth Distance and Maximum Instantaneous Forward Flight Speed in the combined space of frequency and amplitude:

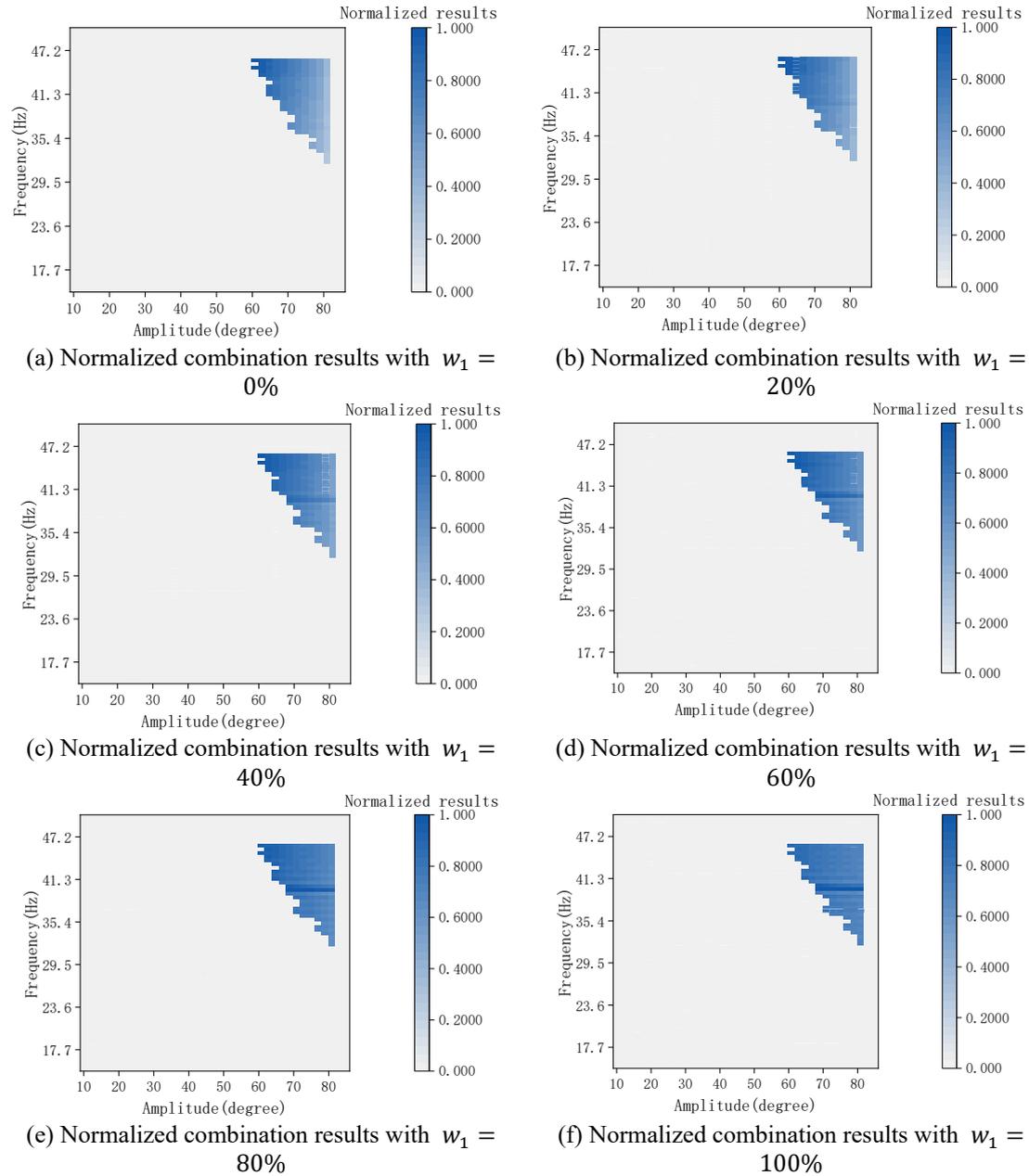

(a) Normalized combination results with $w_1 = 0\%$

(b) Normalized combination results with $w_1 = 20\%$

(c) Normalized combination results with $w_1 = 40\%$

(d) Normalized combination results with $w_1 = 60\%$

(e) Normalized combination results with $w_1 = 80\%$

(f) Normalized combination results with $w_1 = 100\%$

**Fig. 12.** Coupling relationship between the MBSD and the MIFFS in the design space composed of frequency and amplitude with different weights.

To analyze the coupling relationship between Minimal Biorobotic Stealth Distance and Maximum Instantaneous Forward Flight Speed in the combined space of amplitude and wingspan:

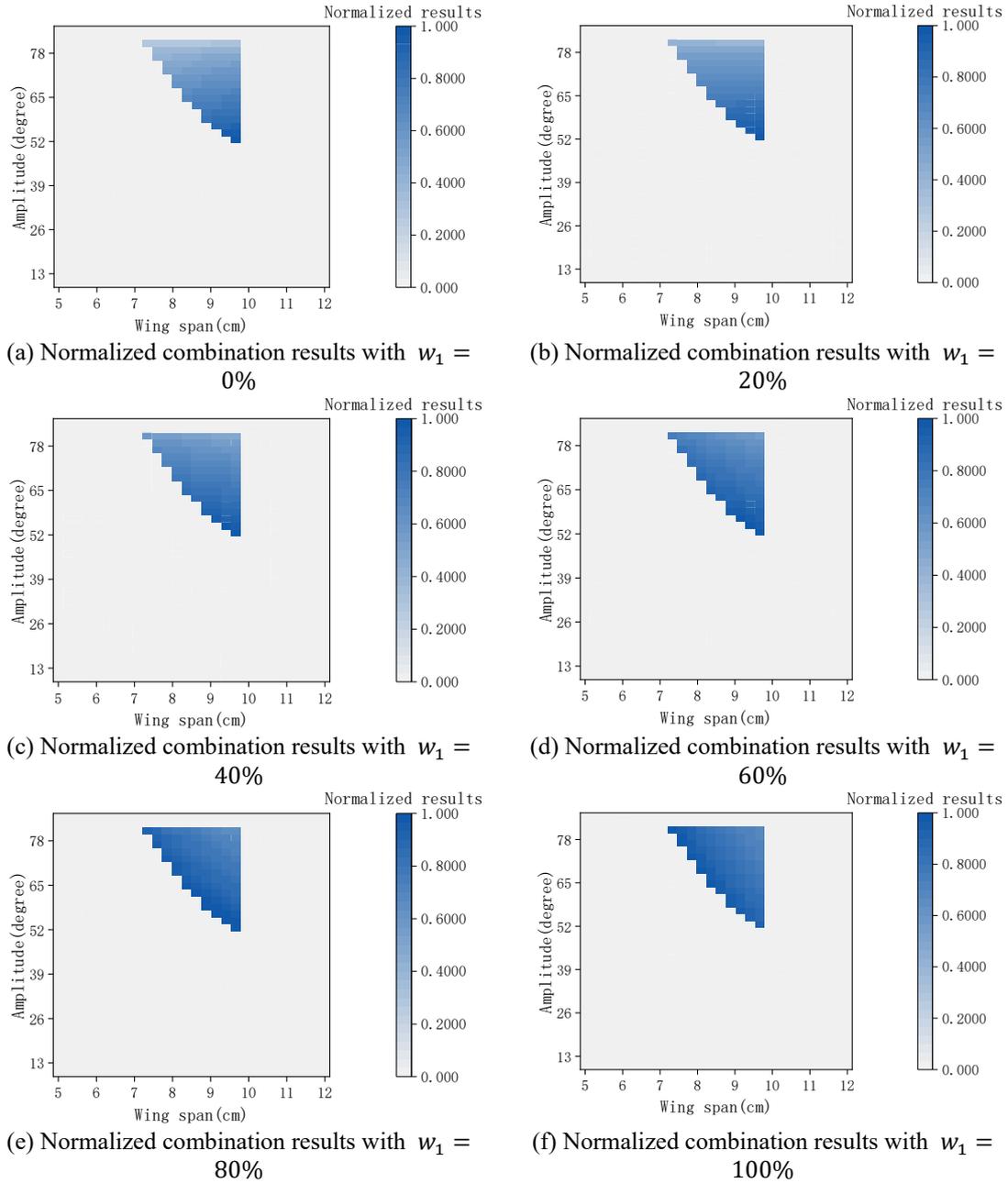

**Fig. 13.** Coupling relationship between the MBSD and the MIFFS in the design space composed of amplitude and wingspan with different weights

Based on Figs 11 to 13, the following can be observed:

1. Compared with the distribution of the MIFFS values, the MIFFS tends to achieve maximum values at high frequencies and large wingspans. In contrast, the inverse value of the MBSD achieves maximum values (i.e., better bionic resemblance) at specific frequencies and small wingspans. This significant difference highlights the conflict between the mechanical attributes (e.g., the MIFFS value) and bionic attributes (e.g., the MBSD value) of biorobotic aircraft under current technological conditions.

2. As shown in Fig. 12, the MIFFS achieves larger values at high frequencies and small

wingspans, while the MBSD achieves larger values at specific frequencies. In the combined space of the MIFFS and the MBSD, aircraft with specific frequencies have a greater design advantage, highlighting the coupling of the two objectives.

3. The trends influenced by wingspan and amplitude show significant similarities for both objectives. Both the inverse value of the MBSD (i.e., better bionic resemblance) and the MIFFS value achieve maximum values at larger wingspans and smaller amplitudes. This suggests that higher bionic resemblance corresponds to higher overall forward flight speed, consistent with existing research[53, 55, 85], highlighting the correlation between the objectives.

4. The MBSD value provides a perspective on balancing the mechanical attributes (e.g., MIFFS value) and bionic attributes (e.g., the MBSD value) of biorobotic aircraft under current technological conditions.

# 6. Analysis of the Coupling Relationship Between Minimal Biorobotic Stealth Distance and Typical Aircraft Indicators from a Multi-Objective Optimization Perspective

## 6.1. Experiment Setup

In the previous chapter, the coupling relationship between Minimal Biorobotic Stealth Distance and typical design parameters was analyzed intuitively. However, this analysis has limitations: 1) Only three key parameters were selected, with the remaining parameters set to their default values, as shown in Table 3.

To further explore the use of the MBSD value in optimization scenarios and investigate the coupling relationships under more parameter conditions for quantitative analysis, this study aims to maximize the aircraft's potential. This study utilizes the following experimental setup for optimization:

1. The SMSEMOA multi-objective algorithm from the Pymoo library[90] is used as the optimization algorithm. The system model established above is employed, with a population size of 200, iterating until convergence.
2. All five design variables are considered.
3. Results that do not meet the constraints are excluded.

## 6.2. Coupling Relationship Between Minimal Biorobotic Stealth Distance and Longest Hover Duration in a Multi-Objective Optimization Context

### 6.2.1. Optimization Result Display

Based on the optimization method described in Section 6.1, all individual samples generated during the optimization process are used for subsequent analysis, as shown in Fig. 14.

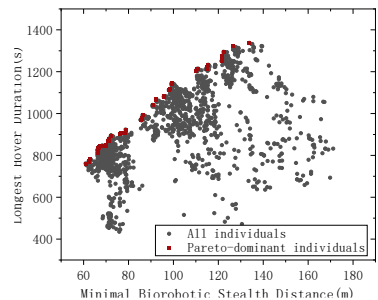

**Fig. 14.** Display of all sample points generated during the optimization process in the design target space.

### 6.2.2. Importance Analysis of Design Parameters Influencing the Coupling Relationship of Two Design Objectives

To analyze the differences in focus between the MBSD and the LHD on design variables, a feature importance analysis is conducted on the obtained samples.

A Random Forest-based feature importance analysis method is used to analyze the importance of the five design variables for the MBSD, with results shown in Fig. 15.

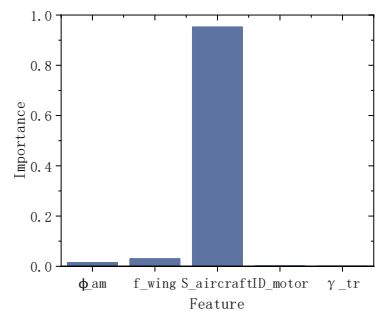

**Fig. 15.** Importance analysis of the MBSD target.

Based on Fig. 15 and the definitions of the MBSD in equations (27) to (29), the following can be observed:

1. $S_{aircraft}$, $f_{wing}$, and $\phi_{am}$ are the most important parameters for the MBSD, consistent with the theoretical relationships defined in the equations.

2. Notably, the wingspan $S_{aircraft}$ has the most significant impact on the MBSD.

A Random Forest-based feature importance analysis method is used to analyze the importance of the five design variables for the LHD, with results shown in Fig. 16.

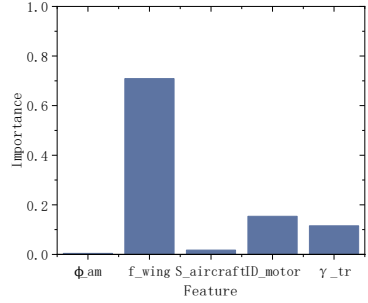

**Fig. 16.** Importance analysis of the LHD target.

Based on Fig. 16, the following can be observed:

1. Consistent with existing research conclusions[5, 46, 78, 80, 81, 86], parameters such as $S_{aircraft}$, $f_{wing}$, $\gamma_{tr}$ and $ID_{motor}$ significantly impact the LHD value of hovering aircraft. This phenomenon also indirectly validates the rationality of the simulation system constructed in Fig. 5, ensuring the effectiveness of subsequent correlation analysis.

2. The impact of amplitude is not significant because the direct-drive biorobotic aircraft adopts a resonant body structure, where the drive system includes springs that resonate with the motor rotor[67, 91, 92], making the system less sensitive to amplitude.

3. The parameters focused on by the LHD partially overlap with those for the MBSD, but not entirely, highlighting the differences in design parameter emphasis between the MBSD and conventional design objectives such as the LHD.

Overall, the differences in focus between the MBSD and common design objectives like the LHD underscore the importance of using the MBSD to balance the mechanical and bionic attributes of biorobotic aircraft under current technological conditions.

### 6.2.3. Correlation Analysis Between Sample Parameters

To further reveal the relationship between the MBSD and the LHD, a correlation analysis is conducted based on the design parameters and target performance of the samples shown in Fig. 14. The analysis results are shown in Fig. 17.

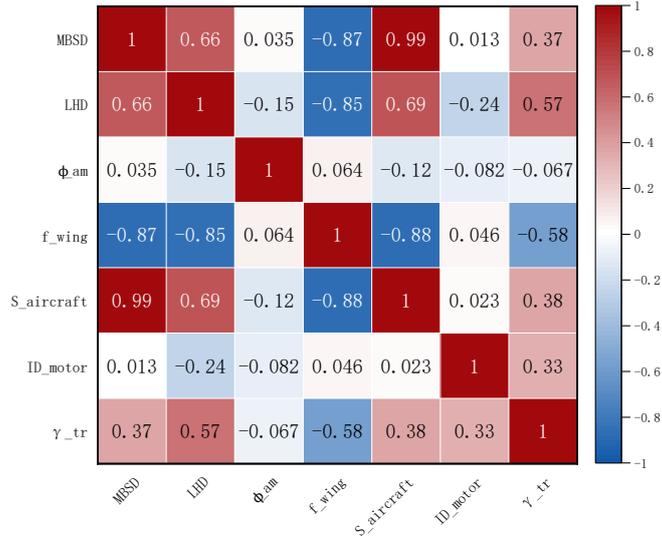

**Fig. 17.** Correlation between the MBSD and the LHD and their respective design variables.

Based on Fig. 17, the following can be observed:

1. Regarding the relationship between design objectives: The correlation coefficient between the LHD and the MBSD is 0.66, indicating a moderate linear relationship between the two objectives. However, since a smaller MBSD value is better (closer to biological resemblance), achieving a longer hover time may require accepting a larger MBSD value, i.e., deviating from the biological reference.
2. The relationship between the MBSD and flapping amplitude $\phi_{am}$ is not significant, but the MBSD is highly influenced by frequency. This is consistent with the definition of the minimum visual resolution distance required to eliminate motion trajectory dissimilarity in equation (29), where amplitude has a relatively smaller impact on trajectory similarity.
3. The MBSD and $S_{aircraft}$ exhibit a strong positive linear correlation (correlation of 0.99), consistent with the characteristics defined in equations (27) to (29).

Overall, the MBSD provides a perspective for balancing the mechanical and bionic attributes of biorobotic aircraft under current technological conditions.

### 6.2.4. Proportional Relationship Between Minimal Biorobotic Stealth Distance and Longest Hover Duration

To further explore the coupling relationship between these two design objectives, the following relationship is defined:

$$E_{LHD} = \frac{LHD}{MQMBSD} \qquad (45)$$

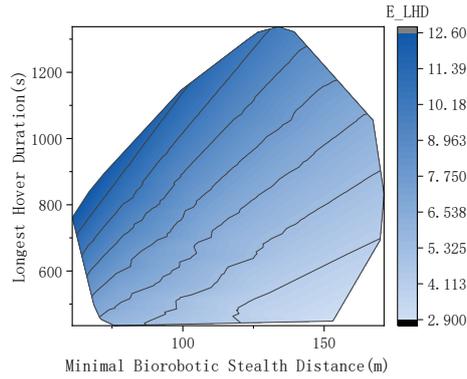

**Fig. 18.** Proportional relationship between the MBSD and the LHD

Based on the definition in equation (45), the samples shown in Fig. 14 are calculated to produce Fig. 18. It can be observed that smaller MBSD values result in the largest $E_{LHD}$, indicating that the more bionic the aircraft, the more sensitive its mechanical performance is to bionic adjustments. This provides a reference for designing biorobotic aircraft: either precisely designing an aircraft with high bionic resemblance or completely abandoning bionic resemblance in favor of mechanical performance to reduce design difficulty. The analysis highlights the importance of the MBSD in balancing the mechanical and bionic attributes of biorobotic aircraft.

## 6.3. Coupling Relationship Between Minimal Biorobotic Stealth Distance and Maximum Instantaneous Forward Flight Speed in a Multi-Objective Optimization Context

### 6.3.1. Optimization Result Display

Based on the optimization method described in Section 6.1, all individual samples generated during the optimization process are used for subsequent analysis, as shown in Fig. 19.

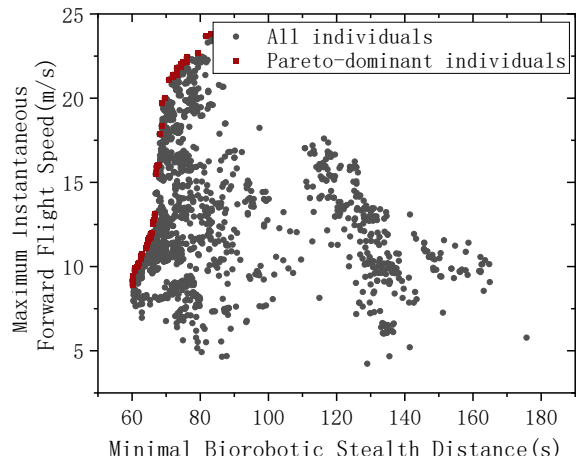

**Fig. 19.** Display of all sample points generated during the optimization process in the design target space.

## 6.3.2. Importance Analysis of Design Parameters Influencing the Coupling Relationship of Two Design Objectives

To analyze the differences in focus between the MBSD and Maximum Instantaneous Forward Flight Speed on design variables, a feature importance analysis is conducted on the obtained samples.

A Random Forest-based feature importance analysis method is used to analyze the importance of the five design variables for the MBSD, with results shown in Fig. 20.

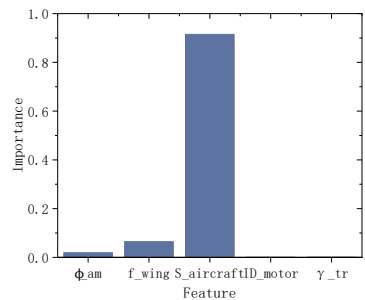

**Fig. 20.** Importance analysis of the MBSD target

Based on Fig. 20 and the definitions of the MBSD in equations (27) to (29), the following can be observed:

1. The distribution is similar to the importance shown in Fig. 15, indicating that the samples obtained from both optimizations are similar. This proves the unbiasedness of the samples.

A Random Forest-based feature importance analysis method is used to analyze the importance of the five design variables for the MIFFS, with results shown in Fig. 21.

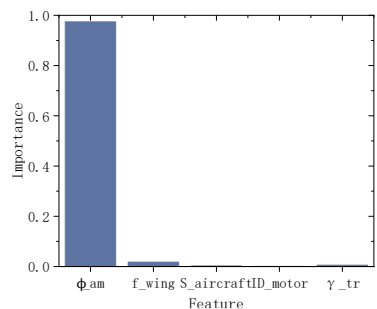

**Fig. 21.** Importance analysis of the MIFFS target

Based on Fig. 21, the following can be observed:

1. The MIFFS value is most influenced by $\phi_{am}$, which differs significantly from the emphasis shown in Fig. 20 for the MBSD.
2. Both the MIFFS and the MBSD are significantly influenced by flapping frequency.

Overall, the differences in focus between the MBSD and common design objectives like the

MIFFS underscore the importance of using the MBSD to balance the mechanical and bionic attributes of biorobotic aircraft under current technological conditions.

### 6.3.3. Correlation Analysis Between Sample Parameters

To further reveal the relationship between the MBSD and the MIFFS, a correlation analysis is conducted based on the design parameters and target performance of the samples shown in Fig. 19. The analysis results are shown in Fig. 22.

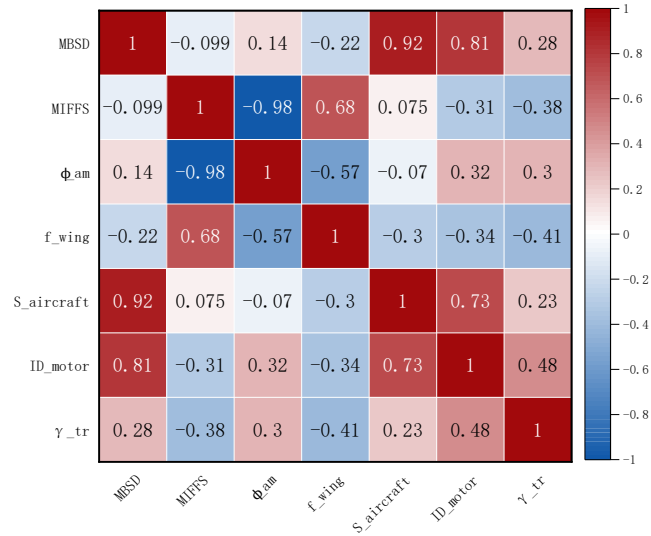

**Fig. 22.** Correlation between the MBSD and the MIFFS and their respective design variables

Based on Fig. 22, the following can be observed:

1. Regarding the relationship between design objectives: the MBSD and the MIFFS are relatively independent but show a certain negative correlation, indicating that achieving better forward flight speed may require parameters that are more similar to those of a biological dragonfly.
2. Regarding design parameters: Under the current conditions, the MBSD and $S_{aircraft}$ still show a strong linear correlation. However, the relationship with $f_{wing}$ and $\phi_{am}$ differs from the previous analysis, likely due to the samples generated during the optimization process (i.e., tending towards maximum forward flight speed).

Overall, the MBSD provides a perspective for balancing the mechanical and bionic attributes of biorobotic aircraft under current technological conditions.

### 6.3.4. Proportional Relationship Between Minimal Biorobotic Stealth Distance and Maximum Instantaneous Forward Flight Speed

To further explore the coupling relationship between these two design objectives, the following relationship is defined. Based on the definition in equation (46), its dimension is time, giving $E_{MIFFS}$ the physical meaning of the time required to traverse the Minimal Biorobotic Stealth Distance.

$$E_{MIFFS} = \frac{MBSD}{MIFFS} \qquad (46)$$

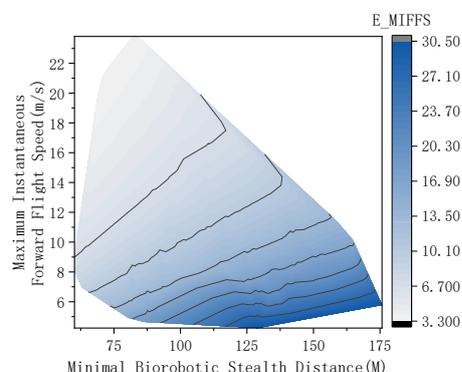

**Fig. 23.** Proportional relationship between the MBSD and the MIFFS

Based on the definition in equation (46), the samples shown in Fig. 19 are calculated to produce Fig. 23. It can be observed that smaller the MBSD values result in the smallest $E_{MIFFS}$ values, with differences approaching an order of magnitude. This indicates that the better the bionic performance of the biorobotic aircraft, the shorter the time required to traverse the Minimal Biorobotic Stealth Distance, thereby minimizing exposure time. This provides a reference for the design of mission scenarios and biorobotic aircraft: biorobotic aircraft may have significant advantages over non-biorobotic aircraft in specific mission scenarios. The analysis highlights the importance of the MBSD in balancing the mechanical and bionic attributes of biorobotic aircraft.

# 7. Integrating Minimal Biorobotic Stealth Distance with Optimization for Typical Mission Scenarios

## 7.1. Experiment Setup

The Minimal Biorobotic Stealth Distance is crucial for enhancing the operational capabilities of biorobotic aircraft. To further demonstrate the practical application of the MBSD value, this study explores its use in a typical biorobotic aircraft mission scenario.

The optimization conducted is single-objective, with the system characterized by non-linearity and discreteness. Therefore, the ISRES algorithm from the Pymoo library[90] was selected for optimization. The population size was set to 400, with a rule of 1/7, gamma of 0.85, and alpha of 0.2, iterating until convergence.

To highlight the MBSD's practical guidance, this section discusses its application in an actual mission scenario.

## 7.2，Composite Objective: Additional Hover Time

To illustrate the impact of the MBSD on mission capability and aircraft design, this study references existing close reconnaissance devices like the PD-100 [93, 94] and sets a mission profile as shown in Fig. 24:

- **Segment 1: Slow Approach:** The operator releases the biorobotic aircraft from a distance, and it slowly moves towards the target area.
- **Segment 2: Circling Reconnaissance:** The aircraft circles the target area at the MBSD radius, conducting reconnaissance.
- **Segment 3: Rapid Approach:** After circling, the aircraft quickly reaches the target using its Maximum Instantaneous Forward Flight Speed.
- **Segment 4: Hovering Reconnaissance:** The aircraft hovers near the target for prolonged reconnaissance.
- **Segment 5: Rapid Retreat:** The aircraft quickly returns to the MBSD radius edge using its Maximum Instantaneous Forward Flight Speed.

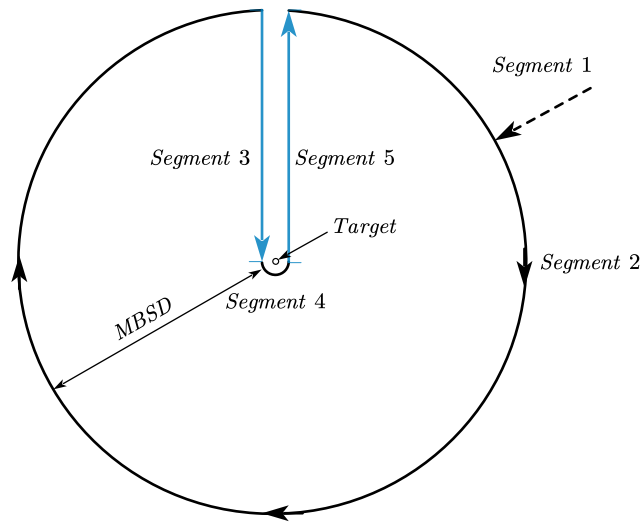

**Fig. 24.** Schematic diagram of a typical mission form

Given the hover capability of the dragonfly-inspired biorobotic aircraft, the following functionalities are realized in the five mission segments:

1. Initiate release from a greater distance with near-hover and minimal forward speed, enhancing tactical flexibility[95].
2. Use hovering in the circling reconnaissance segment to evade detection and increase concealment, improving tactical flexibility[96].
3. Conduct longer reconnaissance in the hovering segment, enhancing tactical

flexibility[97].

Considering the importance of hover capability, a composite indicator, Composite Objective: Additional Hover Time (AHT), is constructed to quantify this feature:

$$t = \frac{E_{total} - \frac{D_i \cdot (2 \cdot \pi + 2)}{V_{front}} \cdot P_{front}}{P_{HOVER}} \quad (47)$$

## 7.3. Optimization Process, Results, and Analysis

Based on the optimization method described, it is evident that a well-optimized design significantly enhances the application potential of biorobotic aircraft, especially those capable of hovering. After 60 generations and 26,401 evaluations, the optimization process converges to a stable objective function value of 763.499 seconds, as shown in Fig. 25.

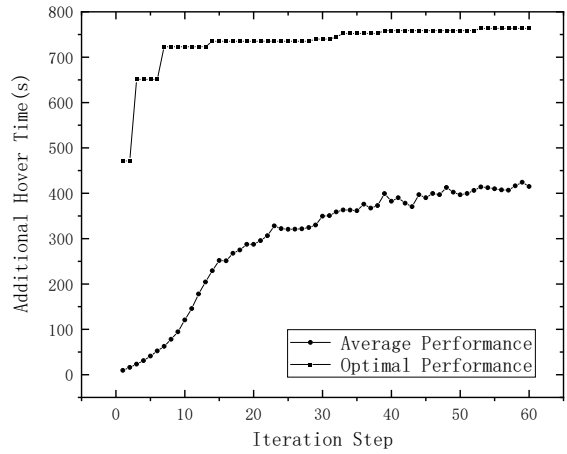

**Fig. 25.** Performance display of sample points generated during the optimization iteration process

The final design parameters achieved an Additional Hover Time of 763.499 seconds after completing the typical mission, as shown in Table 8:

Table 8. Final Optimized Design Parameters

| Design Variable | Optimized Value | Unit |
| --- | --- | --- |
| $S_{aircraft}$ | 83.167 | mm |
| $f_{wing}$ | 29.605 | Hz |
| $\phi_{am}$ | 71.705 | degree |
| $\gamma_{tr}$ | 34.175 | 1 |
| $ID_{motor}$ | 3.780 | / |

Integrating these parameters into the simulation system described in Fig. 5 yields the following typical design target values shown in Table 9:

Table 9. Final Scheme Multi-Objective Performance

| Objective | Value |
|---|---|
| Minimal Biorobotic Stealth Distance | 98.203 |
| Longest Hover Duration | 1111.461 |
| Maximum Instantaneous Forward Flight Speed | 7.773 |

Referring to the relationships between the three objectives and the key design parameters illustrated in Figs. 8 to 13, and the optimization structures shown in Figs. 14 and 19, the final optimized results reflect the characteristics and intent of the MBSD definition. Under current human engineering technology levels, sacrificing some bionic performance (the MBSD value is moderate and not minimal) can achieve better overall mission performance of the aircraft. This demonstrates the MBSD's utility in balancing the mechanical and bionic attributes of biorobotic aircraft under existing technological conditions.

# Conclusion

This paper introduces the Minimal Biorobotic Stealth Distance as a novel quantitative metric to evaluate the bionic resemblance of biorobotic aircraft, specifically focusing on dragonfly-inspired aircrafts. The MBSD metric provides a valuable tool for balancing mechanical performance and bionic characteristics, which is crucial given current technological limitations. Using the DDD-1 dragonfly-inspired aircraft, this study demonstrates the practical application of the MBSD in optimizing aircraft design. The research covers the essential characteristics of MBSD, its coupling relationship with existing performance metrics, and its application in typical mission scenarios.

Experimental results indicate that MBSD correlates well with bionic resemblance and is significantly influenced by design parameters such as wingspan, flapping frequency, and amplitude. Multi-objective optimization reveals important trade-offs between mechanical performance and bionic attributes, providing insights for future biorobotic aircraft design.

**Limitations**

Despite its contributions, this study has several limitations:

1. The evaluation process does not consider the impact of noise, radio frequency, infrared, or olfactory factors, which could affect the stealth and detection of biorobotic aircraft.
2. The construction process does not include the influence of reference objects, which might alter the perception and effectiveness of the MBSD metric.
3. Parameters such as aspect ratio and root-to-tip ratio were not considered, which are important for aerodynamic performance.

4. The study did not take into account the sweep angle, which can significantly affect the flight dynamics and stability of the aircraft.

**Future Work**

Future research should aim to:

1. Further refine the current evaluation metrics by incorporating additional influencing factors such as noise, radio frequency, infrared, and olfactory signatures.
2. Introduce the impact of reference objects into the evaluation process to enhance the accuracy and applicability of the MBSD metric.
3. Consider additional parameters like aspect ratio and root-to-tip ratio to provide a more comprehensive evaluation of biorobotic aircraft performance.
4. Include the sweep angle in the design and optimization process to improve flight dynamics and stability.

By addressing these limitations, the MBSD metric can be further developed to provide a more holistic and effective tool for the design and application of biorobotic aircraft.

# Acknowledgements

Data will be made available on reasonable request. This work is financially supported by "the Fundamental Research Funds for the Central Universities".

# Conflict of interests

The authors declare that they have no known competing financial interests or personal relationships that could have appeared to influence the work reported in this paper

# Appendix A  Wing Inertia Model

To establish the geometric model of the wing, the relevant parameters are defined as follows:

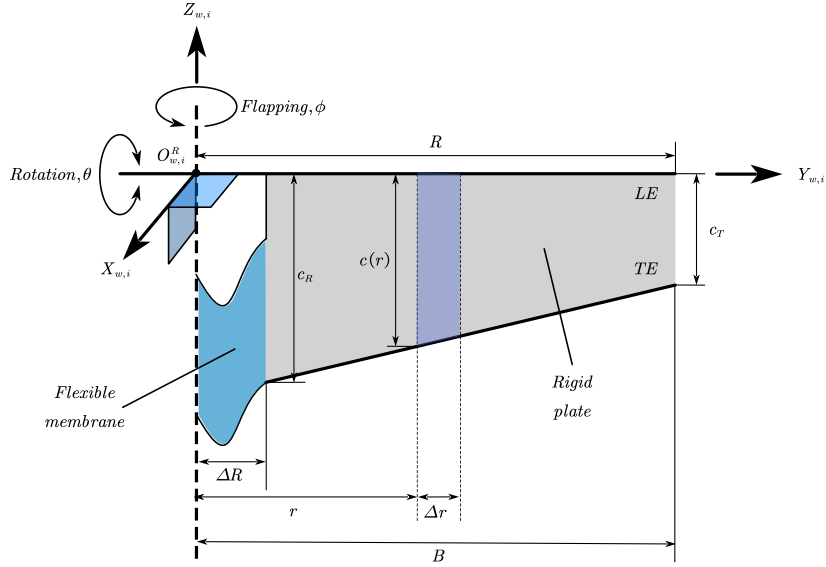

**Fig. 26.** Schematic representation of the geometric parameters and the definition of the wing surface coordinate system for a flapping wing, with the Leading Edge (LE) and the Trailing Edge (TE) clearly delineated.

In the multi-body dynamics simulation, four inertia parameters of the wing are required, defined as:

$$I_{wing} = \begin{bmatrix} J_{xx} & 0 & 0 \\ 0 & J_{yy} & J_{yz} \\ 0 & J_{yz} & J_{zz} \end{bmatrix} \tag{48}$$

Due to the complex geometry of the current wing design, the root chord length, tip chord length, and wing span parameters were determined. Discretization was performed in three directions: thickness, chord, and span, and the overall inertia was estimated using a cuboid approximation method.

To validate the effectiveness of this approach, the inertia parameters obtained from a 3D modeling software were compared with those calculated using the aforementioned method. The wing membrane was assumed to be made of TPU material with a density of 1.1 g/cm³, and a thickness $t$ was assumed.

The wing parameters selected for comparison are in Table 10.:

**Table 10.** Parameters of selected controlled wings

| Parameter | Value (mm) |
|:---:|:---:|
| $R$ | 80.0 |
| $\Delta R$ | 0.0 |
| $c_T$ | 20.0 |
| $c_R$ | 33.3 |
| $t$ | 0.025 |

For discretization, the x-direction was divided into 2 points, the y-direction into 10 points, and the z-direction into 10 points. The results from the 3D modeling software and the calculated model were compared, showing the following relative errors in Table 11:

**Table 11.** Calculation accuracy

| Parameter | 3D Software Value | Calculated Result | Relative Error |
|---|---|---|---|
| $J_{W,XX}$ | $2.674 \times 10^{-7}$ | $2.678 \times 10^{-7}$ | 0.14959% |
| $J_{W,YY}$ | $2.678 \times 10^{-8}$ | $2.688 \times 10^{-8}$ | 0.37341% |
| $J_{W,ZZ}$ | $2.407 \times 10^{-7}$ | $2.409 \times 10^{-7}$ | 0.08309% |
| $J_{W,YZ}$ | $5.574 \times 10^{-8}$ | $5.613 \times 10^{-8}$ | 0.69968% |

# Appendix B  Torsion Spring Model

Given the relatively small weight of the spring in the overall system, this paper does not discuss the dimensional parameters of the spring. Instead, the spring is designed according to the required flapping frequency as follows:

$$K_{rot} = 4\pi^2 f^2 \cdot J_{total} \tag{49}$$

$$J_{total} = J_g + J_w + \gamma_{tr}^2 \cdot J_m \tag{50}$$

where $J_g$, $J_m$, and $J_w$ represents the moment of inertia of the gear, the motor rotor, and the wing. The variable $f$ denotes the desired flapping frequency.

# Appendix C  Viscoelastic Tensioned Membrane Model

This section determines the parameters $C_{wing,i}$ and $K_{wing,i}$ within the model mentioned above. As discussed previously, the flexible membrane part of the wing undergoes reciprocating tension at a relatively high frequency during the flapping process. Consequently, the viscoelastic phenomena[41] that arise must be addressed, necessitating consideration of the loading rate's impact on the tensioning process.

$$T_{VTM,i} = K_{VTM,i} \cdot \theta_i + C_{VTM,i} \cdot \dot{\theta}_i \tag{51}$$

$$K_{VTM,i} = K_{w-s,i} \tag{52}$$

$$C_{VTM,i} = C_{wing,i} \tag{53}$$

In terms of tension condition assessment, this study triggers tension conditions where the absolute values of the torsional angular velocity and the flapping angular velocity are considered, primarily when their motion patterns are inversely related.

$$C_{\text{tens}} = [C_{scale1} \cdot (\dot{\theta}_\iota \cdot \dot{\phi}_\iota) - C_{move}] \tag{54}$$

$$C_{Tension,\ i} = \frac{C_{c1}}{1 + e^{-C_{\text{tens}}{}^2}} \tag{55}$$

Here, $\dot{\theta}_\iota$ represents the rotation angular velocity, and $\dot{\phi}_\iota$ denotes the flapping angular velocity, with $C_{c1}, C_{scale1}$ and $C_{move}$ being constant parameters. To determine the stiffness and damping resulting from the viscoelastic phenomena, presented as follows:

$$K_{wing,i} = C_{Tension,\ i} \cdot K_{w-s,i} \tag{56}$$

$$K_{w-s,i} = \frac{k_{w-s,i}}{k_{w-s-ref,i}} \tag{57}$$

$$k_{w-s,i} = ((k_1 - \frac{1}{\sigma\sqrt{2\pi}} e^{-0.5(\frac{\theta_n - \mu}{\sigma})^8})/k_1)^2 - ((k_1 - \frac{1}{\sigma\sqrt{2\pi}} e^{-0.5(\frac{-\mu}{\sigma})^8})/k_1)^2 \tag{58}$$

$$k_{w-s-ref,i} = ((k_1 - \frac{1}{\sigma\sqrt{2\pi}} e^{-0.5(\frac{\theta_{tat} - \mu}{\sigma})^8})/k_1)^2 - ((k_1 - \frac{1}{\sigma\sqrt{2\pi}} e^{-0.5(\frac{-\mu}{\sigma})^8})/k_1)^2 \tag{59}$$

$$\theta_n = k_1 \cdot \theta \tag{60}$$

$$C_{wing,i} = C_{Tension,\ i} \tag{61}$$

With $\theta_{tat}$, $k_1$, $\sigma$, $\mu$, $C_{wing-ref,i}$ and $C_{Tension,\ i}$ as constant parameters, the model's parameters are selected from a feasible solution set referencing the magnitude of parameters from existing research.

## Appendix D  Quasi Steady Single Wing Model

This section focuses on determining the parameters $T_{y,wing,w,i}$ and $T_{z,wing,w,i}$ within the framework of the proposed model.

The study employs a flapping wing configuration as depicted in Fig. 26. The complete flapping wing is bifurcated into two distinct components: a rigid plate, which executes flapping and rotation movements to generate aerodynamic forces, and a flexible membrane, which, though not contributing to aerodynamic force generation, imparts restraining forces upon being tensioned[46].

This research aims to construct a quasi-steady aerodynamic force model for a single wing, utilizing the quasi-steady aerodynamic estimation model introduced by Ellington and colleagues[98], complemented by the empirical findings from Lee and colleagues[10]. The aerodynamic model primarily encompasses translational aerodynamic forces and moments, rotational aerodynamic forces and moments, and added mass aerodynamic forces and moments[46].

In terms of geometric parameters, a trapezoidal outline, as shown in Fig. 26, has been selected for subsequent simulation endeavors. The flapping motion axis is aligned parallel to the wing root, whereas rotational motion occurs around the leading edge of the flapping wing. The distance from the flapping axis to the wingtip is denoted as $R$, with a wingspan of $B$, and a deviation between the wing root and the flapping axis of $\Delta R = R - B$. The chord lengths at the root and tip are specified as $c_R$ and $c_T$, respectively. For each wing segment $\Delta r$, the chord length as a function of the position $r$ relative to the wing root is defined as $c(r)$.

In the modeling process, the mean chord length ($\bar{c}$), wing area ($S$), aspect ratio ($AR$), and the dimensionless second moment of area ($r_2$) are identified as critical morphological parameters. The definitions of these parameters are as follows[46]:

$$\bar{c} = (c_R + c_T)/2 \tag{62}$$

$$S = \int_{\Delta R}^{R} c(r) dr \tag{63}$$

$$AR = \frac{R^2}{S} \tag{64}$$

$$r_2 = \frac{R_2}{R} = \frac{\sqrt{\frac{1}{S}\int_{\Delta R}^{R} c(r) \cdot r^2 \cdot dr}}{R} \tag{65}$$

In the computation of aerodynamic forces, the most critical parameters are the local angle of attack, $\alpha(r)$, and the local flow velocity, $U(r)$, for each wing segment. These can be calculated using the following formula[99]:

$$\alpha(r) = actan(v_x(r)/v_z(r)) \tag{66}$$

$$U(r) = \sqrt{v_x(r)^2 + v_z(r)^2} \tag{67}$$

where $v_x(r)$ denotes the local velocity component along the x-axis of the wing surface coordinate system, $v_z(r)$ represents the local velocity component along the z-axis of the wing surface coordinate system.

Regarding the calculation of translational aerodynamic forces and moments, which serve as the principal source of aerodynamic forces during the flapping process of the flapping wing, the computational formula is as follows[46]:

$$\Delta F_{x,tr,w,i} = \Delta F_{N,tr} = -1 \cdot C_{N,tr} \cdot 0.5 \cdot \rho \cdot U(r)^2 \cdot c(r) \cdot dr \tag{68}$$

$$\Delta F_{y,tr,w,i} = 0.0 \tag{69}$$

$$\Delta F_{z,tr,w,i} = \Delta F_{T,tr} = C_{T,tr} \cdot 0.5 \cdot \rho \cdot U(r)^2 \cdot c(r) \cdot dr \tag{70}$$

$$C_{N,tr} = [3.48 \cdot \sin(\alpha)] \tag{71}$$

$$C_{T,tr} = [0.4 \cdot \cos(2\alpha)^2] \tag{72}$$

where $\rho$ denotes the air density. The torque generated on each strip is:

$$\begin{bmatrix} \Delta T_{x,tr,i} \\ \Delta T_{y,tr,i} \\ \Delta T_{z,tr,i} \end{bmatrix} = \begin{bmatrix} 0.0 \\ r_{y,tr,w,i} \\ r_{z,tr,w,i} \end{bmatrix} \times \begin{bmatrix} \Delta F_{x,tr,w,i} \\ \Delta F_{y,tr,w,i} \\ \Delta F_{z,tr,w,i} \end{bmatrix} = \begin{bmatrix} +\Delta F_{z,tr,w,i} \cdot L_{S,tr,cop} \\ -\Delta F_{x,tr,w,i} \cdot L_{C,tr,cop} \\ -\Delta F_{x,tr,w,i} \cdot L_{S,tr,cop} \end{bmatrix} \tag{73}$$

$$L_{S,tr,cop} = r \tag{74}$$

$$L_{C,tr,cop} = 0.388 \cdot c(r) \tag{75}$$

The total aerodynamic force and moment on the wing surface, resulting from translational motion within the wing coordinate system, are determined through the following calculation:

$$F_{x,tr,w,i} = \int_{\Delta R}^{R} \Delta F_{x,tr,w,i} \tag{76}$$

$$F_{y,tr,w,i} = 0.0 \tag{77}$$

$$F_{z,tr,w,i} = \int_{\Delta R}^{R} \Delta F_{z,tr,w,i} \tag{78}$$

$$T_{x,tr,i} = \int_{\Delta R}^{R} \Delta T_{x,tr,i} \tag{79}$$

$$T_{y,tr,i} = \int_{\Delta R}^{R} \Delta T_{y,tr,i} \tag{80}$$

$$T_{z,tr,i} = \int_{\Delta R}^{R} \Delta T_{z,tr,i} \tag{81}$$

In calculating rotational aerodynamic forces and moments, these are primarily induced by the significant rotational motion of the flapping wing, attributed to the rotational circulation around the wing. The computational formula is as follows[46]:

$$\Delta F_{x,rot,w,i} = -1 \cdot f_\alpha \cdot f_r \cdot C_{rot,1} \cdot \rho \cdot \dot{\beta}_z \cdot \dot{\beta}_y \cdot r \cdot c(r) \cdot dr \tag{82}$$

$$\Delta F_{x,rot,w,i} = f_\alpha \cdot f_r \cdot C_{rot,1} \cdot \rho \cdot v_x(r) \cdot v_z(r) \cdot c(r) \cdot dr + 2.67 \cdot \rho \cdot \dot{\beta}_y \cdot |\dot{\beta}_y| \\ \cdot \left[ \int_{LE}^{TE} r \cdot x|x| dx \right] \cdot dr \tag{83}$$

$$f_\alpha = \begin{cases} +1, -45° < \alpha < 45° \\ -1, 135° < \alpha < 225° \\ \sqrt{2}\cos(\alpha), otherwise \end{cases} \tag{84}$$

$$\Delta F_{y,rot,w,i} = 0.0 \tag{85}$$

$$\Delta F_{z,rot,w,i} = 0.0 \tag{86}$$

$$C_{rot,1} = 0.842 - 0.507 \cdot Re^{-0.158} \tag{87}$$

$$\dot{\beta}_z = \dot{\phi} \tag{88}$$

$$\dot{\beta}_y = \dot{\theta} \tag{89}$$

where $\dot{\beta}_z$ represents the projection of the current wing surface angular velocity on the $Z_{w,i}$ axis; $\dot{\beta}_y$ denotes the projection of the current wing surface angular velocity on the $Y_{w,i}$ axis; $Re$ is the Reynolds number. The aerodynamic force and moment generated due to rotation are[46]:

$$F_{x,rot,w,i} = \int_{\Delta R}^{R} \Delta F_{x,rot,w,i} \tag{90}$$

$$\begin{bmatrix} T_{x,rot,i} \\ T_{y,rot,i} \\ T_{z,rot,i} \end{bmatrix} = \begin{bmatrix} 0.0 \\ r_{y,rot,w,i} \\ r_{z,rot,w,i} \end{bmatrix} \times \begin{bmatrix} F_{x,rot,w,i} \\ F_{y,rot,w,i} \\ F_{z,rot,w,i} \end{bmatrix} = \begin{bmatrix} 0.0 \\ -\Delta F_{x,rot,w,i} \cdot L_{C,rot,cop} \\ -\Delta F_{x,rot,w,i} \cdot L_{S,rot,cop} \end{bmatrix} \tag{91}$$

$$r_{y,rot,w,i} = 0.993 \cdot R_2 \tag{92}$$

$$r_{z,rot,w,i} = 0.398 \cdot \bar{c} \tag{93}$$

$$F_{x,rot,w,i} = F_{x,rot,w,i} \tag{94}$$

$$F_{y,rot,w,i} = 0.0 \tag{95}$$

$$F_{z,rot,w,i} = 0.0 \tag{96}$$

$$T_{x,rot,i} = 0.0 \tag{97}$$

$$T_{y,rot,i} = -\Delta F_{x,tr,w,i} \cdot L_{C,tr,cop} \tag{98}$$

$$T_{z,rot,i} = -\Delta F_{x,tr,w,i} \cdot L_{S,tr,cop} \tag{99}$$

In calculating added mass aerodynamic forces and moments, these are predominantly induced by the interaction between the flapping wing and the surrounding air during acceleration or deceleration phases. The computational framework is as follows[46]:

$$\Delta F_{x,add,w,i} = -1 \cdot f_{\lambda,\alpha} \cdot f_{AR,\alpha} \cdot f_{a,a} \cdot \frac{\rho\pi}{4} \cdot \ddot{\beta}_z \cdot \sin(|\theta|) \cdot c(r)^2 \cdot r \cdot dr + \ddot{\beta}_y \cdot c(r)^2 \\ \cdot \left[\frac{LE(r) + TE(r)}{2} - rot_x(r)\right] \cdot dr \tag{100}$$

$$f_{\lambda,\alpha} = 47.7 \cdot \lambda^{-0.0019} - 46.7 \tag{101}$$

$$f_{AR,\alpha} = 1.294 - 0.590 \cdot AR^{-0.662} \tag{102}$$

$$f_{a,a} = 0.776 + 1.911 \cdot Re^{-06876} \tag{103}$$

where $\ddot{\beta}_z$ denotes the projection of the current wing surface angular acceleration on the $Z_{w,i}$ axis; $\ddot{\beta}_y$ represents the projection of the current wing surface angular acceleration on the $Y_{w,i}$ axis; $LE(r)$ is the Z-axis coordinate of the leading edge at the position $r$; $TE(r)$ is the Z-axis coordinate of the trailing edge at the position $r$; $rot(r)$ specifies the Z-axis coordinate of the rotation axis at the position $r$.

Consequently, the aerodynamic force and moment generated due to the added mass effect are:

$$F_{x,add,w,i} = \int_{\Delta R}^{R} \Delta F_{x,add,w,i} \tag{104}$$

$$\begin{bmatrix} T_{x,add,i} \\ T_{y,add,i} \\ T_{z,add,i} \end{bmatrix} = \begin{bmatrix} 0.0 \\ r_{y,add,w,i} \\ r_{z,add,w,i} \end{bmatrix} \times \begin{bmatrix} F_{x,add,w,i} \\ F_{y,add,w,i} \\ F_{z,add,w,i} \end{bmatrix} = \begin{bmatrix} 0.0 \\ -\Delta F_{x,add,w,i} \cdot L_{C,add,cop} \\ -\Delta F_{x,add,w,i} \cdot L_{S,add,cop} \end{bmatrix} \tag{105}$$

$$L_{C,add,cop} = 1.078 \cdot R_2 \tag{106}$$

$$L_{S,add,cop} = 0.500 \cdot \bar{c} \tag{107}$$

$$F_{x,add,w,i} = F_{x,add,w,i} \tag{108}$$

$$F_{y,add,w,i} = 0.0 \tag{109}$$

$$F_{z,add,w,i} = 0.0 \tag{110}$$

$$T_{x,add,i} = 0.0 \tag{111}$$

$$T_{y,add,i} = -\Delta F_{x,add,w,i} \cdot L_{C,add,cop} \tag{112}$$

$$T_{z,add,i} = -\Delta F_{x,add,w,i} \cdot L_{S,add,cop} \tag{113}$$

Within the wing surface coordinate system, the wing experiences tri-axial forces and corresponding moments due to aerodynamic phenomena as follows:

$$AF_{x,tol,w,i} = F_{x,tr,w,i} + F_{x,rot,w,i} + F_{x,add,w,i} \tag{114}$$

$$F_{y,tol,w,i} = 0 \tag{115}$$

$$F_{z,tol,w,i} = F_{z,tr,w,i} + F_{z,rot,w,i} + F_{z,add,w,i} \tag{116}$$

$$T_{y,wing,w,i} = T_{y,tr,i} + T_{y,rot,i} + T_{y,add,i} \tag{117}$$

$$T_{z,wing,w,i} = T_{z,tr,i} + T_{z,rot,i} + T_{z,add,i} \tag{118}$$

This section discusses the experimental model based on the water tunnel experiments conducted by Lua et al.[100], using the moth wing as a reference. The motion analyzed includes the flapping motion at the wing root and the twisting motion around the leading edge, following a

specific flapping pattern as outlined and shown in Fig. 27:

$$\phi_{w,i} = \phi_{am} \cdot cos(2\pi \cdot f \cdot t) \tag{119}$$

$$\theta_{w,i} = \theta_{am} \cdot sin(2\pi \cdot f \cdot t) \tag{120}$$

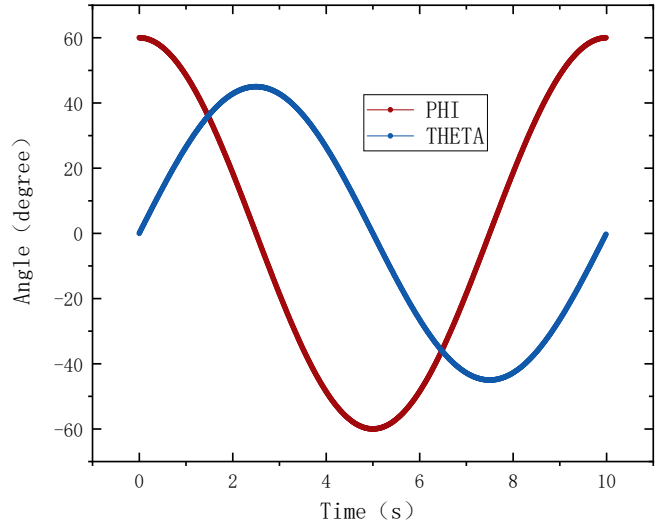

**Fig. 27.** The changing rules of flapping angle and rotation angle selected in the experiment

Given the limitations of the quasi-steady framework, experiments focus on the test rig's validation, selection, and effectiveness of the controller, which revolves around peak accuracy and trend similarity based on the Pearson correlation coefficient. As shown in Table 12 and Fig 28, it is observed that the calculated forces exhibit a trend similar to experimental values, with comparable peak values.

**Table 12.** Verification result statistics

| Conditions | Peak load accuracy | Trend similarity | Average force under QS model | Average force under experiment | Error |
|---|---|---|---|---|---|
| Lift | 6.663% | 0.988 | 0.175 | 0.191 | 8.278% |
| Drag | 6.888% | 0.908 | 0.222 | 0.237 | 6.415% |

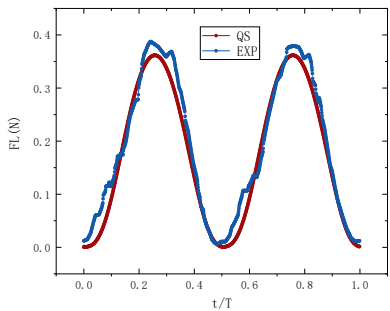 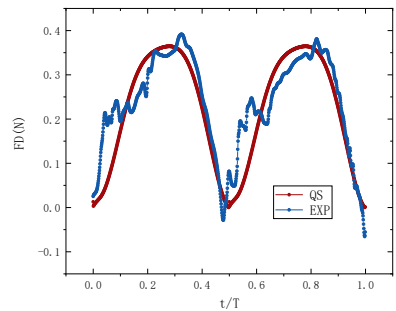

(a) Comparison results for lift    (b) Comparison results for drag

**Fig. 28.** Validation results

# Appendix E  Tandem Wing Correction Influence Model

Drawing upon existing research[47] on aerodynamic interference in tandem wings, where the phase difference is 180 degrees, we extracted lift and drag data to analyze the interference effects on the forewings and hindwings. The analysis shown in the Fig. 29 revealed that:

1. Due to the phase difference of 180 degrees, the data exhibits a specific mirror effect between the forewings and hindwings.
2. The drag of the tandem wings influences both lift and thrust.

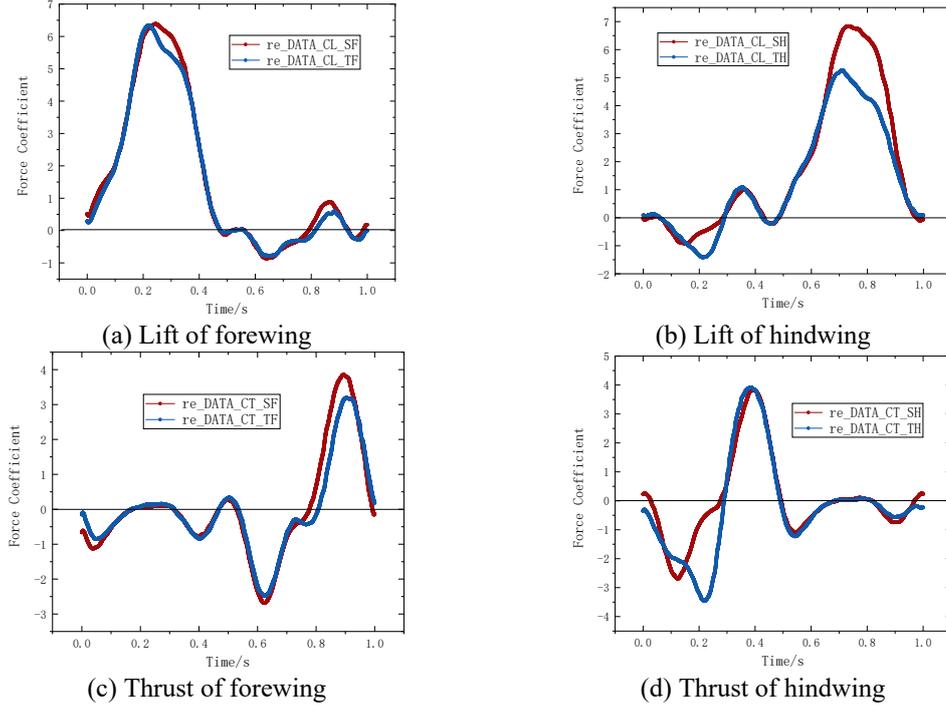

(a) Lift of forewing  (b) Lift of hindwing
(c) Thrust of forewing  (d) Thrust of hindwing
**Fig. 29.**  Raw data for forces with and without tandem wing effects

To further analyze the impact of tandem wing interference on single-wing models, we established the following coefficients.

$$C_T = \frac{F_T - F_S}{F_S} \times 100\% \tag{121}$$

where $F_S$ is the single wing force, $F_T$ is the force under the single wing force under tandem wing configurations.

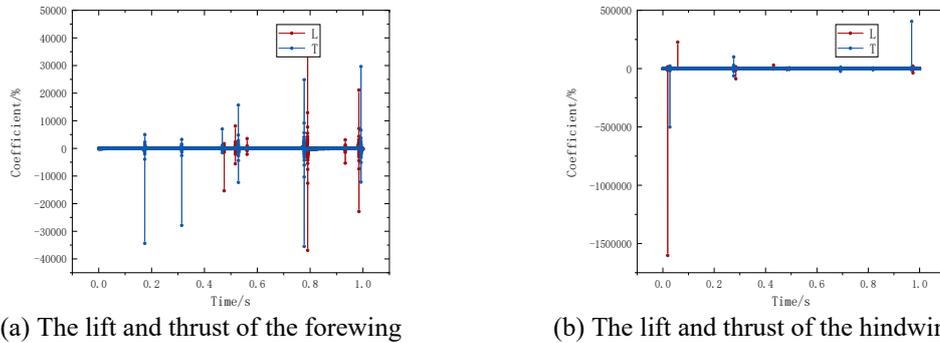

(a) The lift and thrust of the forewing  (b) The lift and thrust of the hindwing
**Fig. 30.**  Date before treatment are jointly affected by the tandem wing interference.

Notably, as shown in Fig.30, regions approaching zero exhibited sharp value increases.

These abrupt changes, resulting from the 1/x relationship, signify rapid ratio increases. To mitigate these anomalies, we utilized the growth gradient as a criterion to exclude these regions. Considering that the aerodynamic forces in these areas are zero or near zero, discussing the ratio of tandem wing aerodynamic interference is irrelevant, thus setting these regions to zero. With a gradient threshold of 0.00001, we obtained data with the abrupt changes removed, as shown in Fig. 31.

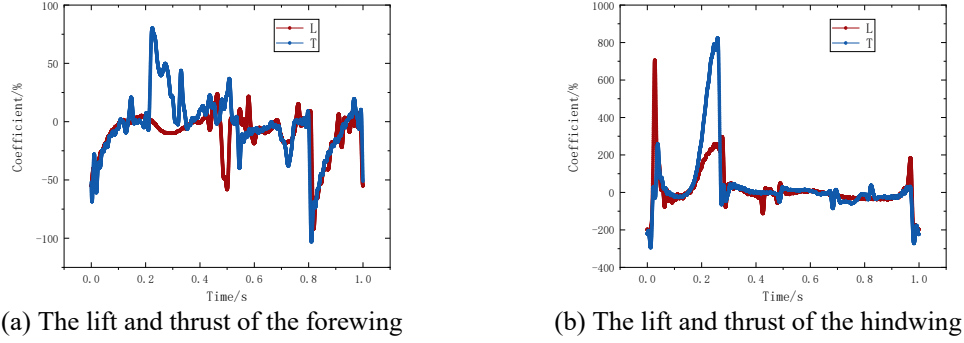

(a) The lift and thrust of the forewing    (b) The lift and thrust of the hindwing

**Fig. 31.** Date after treatments are jointly affected by the tandem wing interference.

Our key findings include:
1. Distinct variation patterns between the forewings and hindwings indicate the need for individualized models to accurately capture the dynamics of tandem wing interference.
2. Despite these differences, the influence of tandem wing interference on the lift and drag characteristics of both wings exhibited similar patterns. This observation led us to hypothesize that a unified model can approximate the effects of tandem wing interference on lift and drag.

Assuming the existence of a general tandem wing interference model, we propose the following formulation:

$$F_T = f(\phi_f, \dot{\phi}_f, \phi_h, \dot{\phi}_h) \cdot F_S \tag{122}$$

This model, inspired by existing aerodynamic frameworks, is further refined to:

$$F_T = f(X_0, X_1, X_2, X_3, X_4, X_5, X_6, X_7, X_8, X_9, X_{10}, X_{11}) \cdot F_S \tag{123}$$

$$X_0 = \dot{\phi}_f/(w_{maxf} \cdot C_{am}) \tag{124}$$

$$X_1 = \dot{\phi}_h/(w_{maxh} \cdot C_{am}) \tag{125}$$

$$X_2 = (\dot{\phi}_f - \dot{\phi}_h)/(w_{maxd} \cdot C_{am}) \tag{126}$$

$$X_3 = \phi_f/C_{am} \tag{127}$$

$$X_5 = (\phi_f - \phi_h)/C_{am} \tag{128}$$

$$X_6 = \sin(2 \cdot \phi_f \cdot (\pi/C_{am})) \tag{129}$$

$$X_7 = \sin(2 \cdot \phi_h \cdot (\pi/C_{am})) \tag{130}$$

$$X_8 = \sin(4 \cdot \phi_f \cdot (\pi/C_{am})) \tag{131}$$

$$X_9 = \sin(4 \cdot \phi_h \cdot (\pi/C_{am})) \tag{132}$$

$$X_{10} = \sin(8 \cdot \phi_f \cdot (\pi/C_{am})) \tag{133}$$

$$X_{11} = \sin(8 \cdot \phi_h \cdot (\pi/C_{am})) \tag{134}$$

Utilizing data presented in Fig. 29, we employed symbolic regression to construct interference models for both fore and hindwings. The iterative process of symbolic regression demonstrated gradual convergence, as shown in Fig. 32, highlighting the effectiveness of this method in identifying underlying aerodynamic relationships.

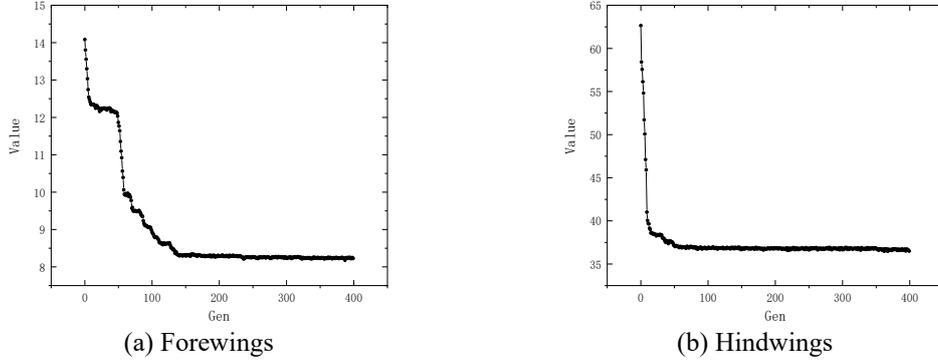

(a) Forewings  (b) Hindwings

**Fig. 32.** Iterative process of symbolic regression

The symbolic regression analysis yielded the following interference coefficients for the forewing:

$$\begin{aligned} C_{TF} = &\; X_0 \cdot X_8 \cdot (-11.453 \cdot X_0 \cdot (-5 \cdot X_6 + X_7) - 22.906 \cdot X_1 + 11.453 \cdot X_2 \\ &- 11.453 \cdot X_4 + 11.453 \cdot X_5 - 11.453 \cdot X_7 - 11.453 \\ &\cdot \sin(\sin(X_8))) + X_0 \cdot X_9 \cdot (11.453 \cdot X_{10} + 11.453 \cdot X_{11} + 11.453 \\ &\cdot X_2 - 22.906 \cdot X_4 - 11.453 \cdot X_5 \cdot (X_1 - X_{11} - X_5) - 11.453 \cdot X_7) \\ &+ 9 \cdot X_1 - 7.892 \cdot X_2 + 7.892 \cdot X_4 + X_8 - 6.166 \end{aligned} \tag{135}$$

The symbolic regression analysis yielded the following interference coefficients for the hindwing:

$$\begin{aligned} C_{TH} = &\; 49.314 \cdot X_1 - (X_2 - X_5 - \sin(X_1 - 124.935)) \cdot (X_2 + X_4 - X_8 - 0.994) \\ &\cdot (X_3 + X_4 + X_6 + X_7) \cdot (-2X_1 - X_{10} - X_{11} + X_2 - X_7 \cdot (X_1 + X_7 \\ &- \sin(X_{10} - X_5)) \cdot (X_1 + X_{11} - X_9 - 45.822) - X_8 - (X_2 + X_4 \\ &- \sin(X_1 - 124.935))(2 \cdot X_0 - 3 \cdot X_{10} - X_9 - 34.855)(X_3 + X_4 \\ &+ X_6 + X_7) - 49.314) \end{aligned} \tag{136}$$

A comparative analysis between the symbolic regression results and original data, as shown in Fig. 33, revealed a correlation coefficient of 0.88037 for the forewing and 0.88618 for the hindwing, relative to the average impact of interference. These coefficients validate the robustness of our proposed models in capturing the nuanced effects of tandem wing interference.

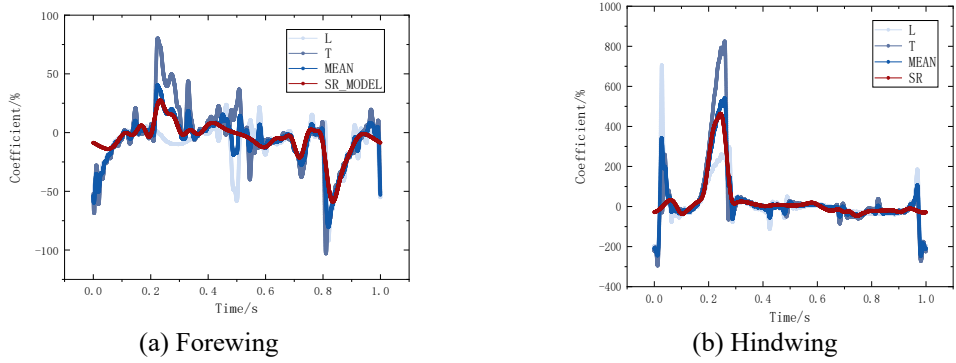

(a) Forewing        (b) Hindwing

**Fig. 33.** Comparison of the construction results of the tandem wing effect on the wing based on

# Appendix F  Motor Parameter Database

For the power characteristic analysis mentioned above, the typical available motor models and key parameters for hovercraft with a take-off weight of approximately 100g are shown in Table 13:

**Table 13.** Motor data selected in this paper

| Motor | Rated Voltage (V) | Max Current (A) | $I_0(mA)$ | $R_0(\Omega)$ | $K_v(rpm/V)$ | Weight (g) |
|---|---|---|---|---|---|---|
| ECX-Prime-235- -6V | 6 | 2.04 | 83.8 | 2.94 | 6310 | 3 |
| ECX-Prime-235- -12V | 12 | 1.02 | 41.9 | 11.7 | 3150 | 3 |
| CN-174-3V | 3 | 3.92 | 149 | 0.766 | 25800 | 3 |
| CN-174-6V | 6 | 1.72 | 58.8 | 3.49 | 10800 | 3 |
| CN-174-12V | 12 | 0.97 | 29.8 | 12.4 | 5460 | 3 |
| CN-173-6V | 6 | 0.688 | 46.5 | 8.72 | 7900 | 3 |
| CN-173-12V | 12 | 0.188 | 16.2 | 63.8 | 3040 | 3 |
| CN-176-6V | 6 | 3.34 | 128 | 1.8 | 6160 | 6 |
| CN-176-9V | 9 | 1.7 | 63.4 | 5.3 | 3360 | 6 |
| CN-176-12V | 12 | 1.43 | 50.9 | 8.38 | 2640 | 6 |
| CN-175-6 | 6 | 1.98 | 105 | 3.02 | 6230 | 6 |
| CN-175-12 | 12 | 1.54 | 69 | 7.8 | 3780 | 6 |
| CN-175-24 | 24 | 0.755 | 33.2 | 31.8 | 1840 | 6 |
| CN_0620_B_FMM-6V | 6 | 0.79788 | 56 | 8.8 | 8761 | 2.5 |
| CN_0620_B_FMM-12V | 12 | 1.55382 | 18 | 60.2 | 3386 | 2.5 |
| CN_0824_B_FMM-6V | 6 | 5.248 | 55 | 2.91 | 5968 | 5.2 |
| CN_0824_B_FMM-12V | 12 | 10.02 | 31 | 10.7 | 3183 | 5.2 |
| otecs0921w-3 | 3 | 0.925 | 45 | 3.24 | 5606 | 6.5 |
| otecs0921w-6 | 6 | 1.99 | 69 | 3.02 | 6182 | 6.5 |
| otecs0921w-12 | 12 | 1.685 | 75 | 7.12 | 3663 | 6.5 |
| otecs0921w-24 | 24 | 0.759 | 22 | 31.6 | 1830 | 6.5 |

# Appendix G  Cooling System Model

Referring to existing studies, the thermal resistance of natural convection is defined as follows[101]:

$$Rth_{conv} = \frac{1}{h_c \cdot S_i} \tag{137}$$

where $h_c$ is the natural convection heat transfer coefficient, and $S_i$ is the heat dissipation area of the given motor. To calculate $h_c$, data from reference[102] is used, assuming the heat dissipation system is arranged similarly to the configuration provided in the reference. Thus, the thermal resistance is:

$$Rth_i = Rth_{stand} \frac{S_{stand}}{S_i} \tag{138}$$

where $R_{stand}$ is 9.88 K/W and $S_{stand}$ is the external surface area of the motor. The required heat dissipation power is then calculated as:

$$P_{HD} = \frac{\Delta T}{Rth_i} \tag{139}$$

where $\Delta T$ is the temperature difference and $P_{HD}$ is the maximum required heat dissipation power.

# Appendix H  Trajectory Generator

To tackle this issue, we employ a motion strategy organized around the concept of Central Pattern Generators (CPGs)[103] to generate the desired trajectory. The equation (1) describes the expected motion of each wing.

$$\phi_{exp,i}^n = A_i \cdot \sin(2\pi f \cdot t + \varphi_i) \tag{140}$$

where $\phi_{exp,i}^n$ denotes the expected flapping angle position of the i-th wing for the next n steps, $A_i$ denotes the desired flapping amplitude for the *i*-th wing, $f$ represents the flapping frequency, and $\varphi_i$ signifies the phase difference of the *i*-th wing. The experimental Settings section provides the parameters used for this study.

# Appendix I  PID controller

The expression for the PID controller can be represented as follows:

$$T_{M,i} = K_p \cdot e_i(t) + K_i \cdot \int e_i(t)\,dt + K_d \cdot \frac{de_i(t)}{dt} \tag{141}$$

where $K_p$, $K_i$, and $K_d$ are the parameters of the PID controller, and $e_i(t)$ is the tracking error for the $i$-th wing.

Based on existing research, the PID parameters significantly impact the results. Therefore, the PID values were set as shown in the Table 14. These values were maintained for all experiments except in Section 5.5, where the adaptability of the combined controller is discussed.

**Table 14.** PID parameter description

| Parameter | Value |
|---|---|
| $K_p$ | $4.80 \times 10^{-1}$ |
| $K_i$ | $2.00 \times 10^{-5}$ |
| $K_d$ | $7.00 \times 10^{-4}$ |

# Appendix J  First-order Motor and Transmission System Model

In direct-drive aircraft, the motor is a critical component. The wing speed ($w_{wing}$) and torque ($T_{wing}$) can be calculated based on the given trajectory and using the aforementioned aerodynamic model. To analyze the energy for friction heat in the transmission system, a model[65] can be represented as follows:

$$w_m = w_{wing} \cdot \gamma_{tr} \tag{142}$$

$$T_m = \frac{T_{wing}}{\gamma_{tr} \cdot \eta_{tr}} \tag{143}$$

where $\gamma_{tr}$ is the transmission ratio; $\eta_{tr}$ is the transmission efficiency.

In the aspect of the model that can analyze the energy characteristic of the actuator system, the first-order motor model[104, 105] is established as follows:

$$U_m = R_0 I_m + \frac{w_m}{K_{v-rad}} \tag{144}$$

$$I_m = I_0 + \frac{T_m}{K_t} \tag{145}$$

$$P_m = U_m I_m \tag{146}$$

$$K_{v-rad} = \frac{60}{2\pi} K_v \tag{147}$$

$$K_t = \frac{30}{\pi \cdot K_v} \tag{148}$$

where $R_0$ is the motor internal resistance; $I_0$ is the motor no-load current; $K_v$ is the motor velocity constant.

Due to the high-frequency reciprocating motion required for flapping in direct-drive aircraft, in-runner brushless motors are often used. These motors are relatively few in number, so the design process utilizes motor parameters from Appendix F, rather than using estimation models, to obtain the values for $R_0$, $I_0$, and $K_v$.

## Appendix K  Forward Flight Area Estimation Module

The forward flight area of the aircraft primarily consists of the area covered by the four wings and two sets of drive systems. The calculations are as follows:

$$S_w = \frac{C_T + C_R}{2}(R - \Delta R) \tag{149}$$

$$S_{act} = C_R \cdot [(\gamma_{tr} + 1) \cdot m_{tr} \cdot t_{motor}] \tag{150}$$

$$S_{front-H} = 4 \cdot S_w + 2 \cdot S_{act} \tag{151}$$

where $m_{tr}$ is the modulus of the transmission system; $t_{motor}$ is the number of teeth on the motor gear.

## Appendix L  Battery and Electronic System Model

Referring to existing studies[67, 91, 106], the battery model is constructed as follows:

$$M_{bat} = \eta_{bat} \cdot M_{TF} \tag{152}$$

$$E_{bat} = \rho_{ebat} \cdot M_{bat} \cdot \eta_{\text{Boost}} \cdot \eta_{used} \tag{153}$$

where $\rho_{ebat}$ is the energy density; $\eta_{bat}$ is the proportion of the battery to the take-off weight; $\eta_{used}$ is the usable capacity of the battery; $\eta_{\text{Boost}}$ is the efficiency of the boost circuit.

For the initial design phase of the aircraft, this paper primarily considers the energy characteristics of the aircraft. Therefore, the power characteristics of the onboard electronic system are given by:

$$P_{other} = K_{elc} \tag{154}$$

where $K_{elc}$ is selected based on existing aircraft data.